\documentclass[12pt,noinfoline]{imsart}

\usepackage{natbib} 
\usepackage{mathtools} 
\usepackage{booktabs} 
\usepackage{tikz} 

\usepackage{geometry}

\usepackage{url,hyperref,lineno,microtype,subcaption}
\usepackage[utf8]{inputenc}
\usepackage{amsmath,amsfonts,amssymb}
\usepackage{scrextend}
\usepackage{graphicx}
\usepackage{listings}
\usepackage{bbm}
\usepackage{xcolor}
\usepackage{amsthm}
\usepackage{bbm}
\usepackage{xcolor}
\usepackage{comment}
\usepackage[textsize=tiny]{todonotes}
\usepackage{listings}
\usepackage{color}
\usepackage{xcolor}
\usepackage{listofitems}
\usepackage{etoolbox}
\usepackage{mwe}
\usepackage{graphics}
\usepackage{graphicx}
\usepackage{algorithm}
\usepackage{algpseudocode}
\usepackage{natbib}
\usepackage{hyperref,nameref}
 \usepackage{tikz}
\usetikzlibrary{matrix}
\usepackage{graphicx}
\usepackage{diagbox}
\usepackage{amsmath}
\usepackage{amsfonts}
\usepackage{relsize}
\everymath{\displaystyle}
\usepackage[normalem]{ulem}

\usepackage{todonotes} 

\newcommand{\F}[1]{\mathcal{F}_{#1}} 
\newcommand{\B}{\mathcal{B}} 
\newcommand{\fiber}[1]{\mathcal{F}_{#1}} 

\newcommand{\St}{\mathcal{S}} 
\newcommand{\At}{\mathcal{A}} 
\newcommand{\Qt}{p} 
\newcommand{\EXP}{\mathbb{E}}
\newcommand{\filtration}[1]{\mathcal{I}_{#1}}

\newcommand{\norm}[2]{|| #1 ||_{#2}}

\newcommand{\criticparam}[1]{\omega_{#1}}
\newcommand{\actorparam}[1]{\theta_{#1}}
\newcommand{\actorgradestim}{\hat{g_a}}
\newcommand{\criticgradestim}{\hat{g_c}}
\newcommand{\actorgrad}{g_a}
\newcommand{\criticgrad}{g_c}
\newcommand{\innerProd}[3]{\langle #1, #2\rangle_{#3}}

\newcommand{\subf}[2]{%
  {\small\begin{tabular}[t]{@{}c@{}}
  #1\\#2
  \end{tabular}}%
}

\definecolor{dkgreen}{rgb}{0,0.6,0}
\definecolor{gray}{rgb}{0.5,0.5,0.5}
\definecolor{mauve}{rgb}{0.58,0,0.82}

\newtheorem{theorem}{Theorem}[section]
\newtheorem{lemma}{Lemma}[section]

\newtheorem{corollary}{Corollary}[section]
\newtheorem{claim}{Claim}[section]

\newtheorem{assumption}{Assumption}[section]

\newtheorem{definition}{Definition}[section]

\newtheorem{condition}{Condition}[section]

\tikzset{>=latex} 
\colorlet{myred}{red!80!black}
\colorlet{myblue}{blue!80!black}
\colorlet{mygreen}{green!60!black}
\colorlet{mydarkred}{myred!40!black}
\colorlet{mydarkblue}{myblue!40!black}
\colorlet{mydarkgreen}{mygreen!40!black}
\tikzstyle{node}=[very thick,circle,draw=myblue,minimum size=22,inner sep=0.5,outer sep=0.6]
\tikzstyle{connect}=[->,thick,mydarkblue,shorten >=1]
\tikzset{ 
  node 1/.style={node,mydarkgreen,draw=mygreen,fill=mygreen!25},
  node 2/.style={node,mydarkblue,draw=myblue,fill=myblue!20},
  node 3/.style={node,mydarkred,draw=myred,fill=myred!20},
}

\begin{document}

\title{Learning to Sample Fibers for Goodness-of-Fit Testing}

\author{Ivan Gvozdanovi\'c and Sonja Petrovi\'c}\thanksref{addr1}\ead[label=e3]{sonja.petrovic@illinoistech.edu}

\thankstext{addr1}{Department of Applied Mathematics, Illinois Institute of Technology.}
\begin{abstract}


We consider the problem of  constructing exact goodness-of-fit tests for discrete exponential family models. This classical problem remains practically unsolved for many types of structured or sparse data, as it rests on a computationally difficult core task: to produce a reliable sample from lattice points in a high-dimensional polytope. We translate the problem into a Markov decision process and demonstrate a reinforcement learning approach for learning `good moves'  for sampling. We illustrate the approach on data sets and models for which traditional MCMC samplers converge too slowly due to problem size, sparsity structure, and the requirement to use prohibitive non-linear algebra computations in the process. The differentiating factor is the use of scalable tools from \emph{linear} algebra in the context of theoretical guarantees provided by \emph{non-linear} algebra.  Our algorithm is based on an actor-critic sampling scheme, with provable convergence. 
 The discovered moves can be used to efficiently obtain an exchangeable sample, significantly cutting computational times with regards to statistical testing. 

\end{abstract}

\maketitle

%

\section{Introduction} 

Given a categorical data vector $u\in\mathbb Z^d$, and a statistical model $\mathcal M$, a goodness-of-fit test determines whether the model $\mathcal M$ is a suitable explanation for  the data $u$. Determining model fit is a fundamental task in statistical inference; lack thereof means  that parameter estimate interpretations are simply meaningless. Contrary to popular intuition, computational tools for assessing model accuracy have not kept up with the explosion of the types of models for categorical data, which include relational data and networks. Such a state of affairs leaves practitioners to interpreting results from models that have not been validated adequately; see \cite{Johnson-BayesChiSq}. 

We consider log-affine models for categorical data,  as defined in \cite{Lauritzen}, which are exponential family \citep{Barndorff-Nielsen} models  for discrete random vectors with finite state spaces: 
$$f_\theta(u)= h(u) e^{\eta(\theta)^T t(u)-\psi(\theta)}.$$
Here, the sufficient {statistic} is $t(u):\mathcal U\to\mathbb R^m$ , {natural parameter} $\eta(\theta):\Theta\to\mathbb R^m$, and log-partition function   $\psi:\Theta\to\mathbb R$.\footnote{See Appendix~\ref{appendix: log-linear models} for more details on the model family.} 
Typical examples include networks or relational data and  contingency tables  with structural zeros \cite{FienbergWasserman1981categorical,Dob2012};  and first- and higher-order Markov chains \citep{BesagMondal}. 
\cite{goldenberg2010survey}  
illustrates that there is no standard basis for model comparison and assessing goodness of fit. We seek a scalable exact testing method that can be applied across model families with different structures.

An exact goodness-of-fit test for all such models can be theoretically constructed, provided the existence of a scalable algorithm for sampling from the conditional distributions given sufficient statistics of the model. 
For  log-affine models, the support set for this conditional distribution is the set of lattice points in a polytope \citep{DS98}. Namely,  considering the contingency table representation of $u$ allows for marginals to be computed by $t(u)=Mu$ for some integer matrix $M\in\mathbb Z^{n\times d}$ of suitable size.  The matrix  $M$ is known as \emph{the design matrix} of the discrete exponential family.  Define the polytope $\mathcal P_M(b):=\{x\in\mathbb R^d: Mx=b, x\geq 0\}$. Following  algebraic statistics literature \citep{DSS09}, the set of integer points $\mathbb Z^d\cap\mathcal P_M(b)$ is called the $b$-\emph{fiber} of $M$, and denoted by $\mathcal{F}_M(b)$. 
Due to its universality, it is well understood that  exploration of lattice points in a polytope is an NP-hard problem. 

For most categorical datasets  in practice,  the sheer problem size or sparsity  render all available computational tools practically infeasible. Since for prohibitively large and high-dimensional fibers, the challenge is computational, we solve the  fiber sampling problem using reinforcement learning. Specifically, the present paper solves the following:

{\bf Problem.}  \emph{ Given a log-affine model with the design matrix $M\in\mathbb Z^{n\times d}$ and an observed categorical data vector $u\in\mathbb Z^d_{\geq 0}$, produce a uniform sample from the fiber $\mathcal{F}_M(Mu)$.
} 

Theory of algebraic statistics, which we outline in later sections, solves this problem using a set of vectors that, in principle, can be computed from $M$. In practice, this core computation fails for $d$ above $\sim 50$. In addition, the theory does not apply directly to data that are sparse or models that contain structural zeros. We demonstrate our solution on data sets with these challenges.
The paper is organized as follows. \begin{itemize}
\item  We formulate the fiber sampling problem as a Markov decision process in \S~\ref{section:setup}. 
\item   Our actor-critic fiber sampling scheme is presented in \S~\ref{sec: algorithm}. It builds on  
 the existing A2C algorithm proposed by \cite{mnih2016asynchronous}.  
Our modified version  uses the same feature extractor for both  actor and  critic networks, hence reducing the number of parameters, as well as a masking module for the actor network which reduces the $\ell_1$ norm of the generated move, allowing the agent to move through sparse fibers.
\item  We provide the proof of asymptotic convergence to the optimal policy in \S~\ref{section: convergence_proof}. 
  To the best of our knowledge, this is the first proof of convergence for an actor-critic algorithm using (1) a generalized advantage estimator  introduced in \cite{schulman2015high},  and (2)  the shared feature extractor.  
Proofs are delegated to the appendix. 
  \item Section~\ref{sec: simulations} contains experiments on real and synthetic data that appear in the literature. We demonstrate  that the algorithm can: 1) reproduce valid statistical results, 2) scale on dense and sparse fibers, and 3) handle structural zeros in the model, which is famously impossible, even with a Markov basis. Regarding 3), note that one needs -- provably \cite[Proposition 2.1]{HT:10}  -- an even larger set of actions to move on fibers with structural zeros. 
\end{itemize}

\paragraph{Interpretation of results.} The optimal policy for fiber sampling gives a set of actions (moves) that connect the fiber, that is designed in a way to maximize the expected reward. The reward function is set to penalize escaping the polytope, thus the agent is rewarded for fiber discovery from any point. In goodness-of-fit testing, the optimal policy is used to sample points in the fiber of the model given the sufficient statsitics of the observed data. The target distribution for this sample can be changed by way of an acceptance ratio similar to the Metropolis-Hastings algorithm. For our simulations, we perform the \emph{conditional volume test} \cite{DiaconisEfron}, producing a sample from the uniform distribution on the fiber.  The samples we construct are exchangeable; their construction follows \cite{BesagCliffordExchangableSample}, who  argue for its superiority over any single Markov chain sample on the fiber. 

\paragraph{Practical considerations.}  In order for the agent to be able to move through a very high-dimensional and potentially sparse fiber, we do not require the use of any nonlinear algebra or combinatorics software for bases determination.  In addition, we also discuss an optional additional step to decompose the initial solution and allow the agent to learn on sub-problems. This decomposition step further alleviates  the computational burden of producing moves for large $M$.

\paragraph{Related literature.} 
We draw inspiration from  two kinds of sources. One is algebraic statistics, namely \cite{HAT12} and    \cite{RUMBA}, who successfully use the  \emph{linear-algebra} object -- a lattice basis of $M$ -- as a starting set of moves, and build fiber samplers using it. In particular  \cite{RUMBA} creates a biased sampling scheme in its random exploration in order to guide the search towards denser regions of the fiber; it requires careful parameter tuning. 
The second inspiration comes from the use of machine learning in combinatorial optimization. For example,  \cite{RLforIP-cut2020}   presents an RL approach to the cutting plane method used in all modern integer programming solvers. A thorough overview and perspective of applicability and impact of RL in combinatorial optimization is provided in \cite{MLforCombinOptim-Bengio2021}. 
Very recently, \cite{wu2022turning}  proposed casting a famous NP-complete problem into a Markov Decision Process and applying a specific reinforcement learning algorithm to solve it: they focus on the general integer feasibility problem, whether a linear system of equations and inequalities has an integer solution.  In fact, \cite{wu2022turning} served as our the starting point in terms of algorithm design; in  Appendix~\ref{appendix:DeLoera}, we draw parallels and highlight differences in our work.

\subsection{Historical context for learning to sample} 
A brief look back into the history of sampling discrete structures suggests that randomized search algorithms could simultaneously tackle sparsity, dimensionality, and time complexity of the problem, given their 
  long history of successfully solving NP-hard problems. The  Monte Carlo method \cite{MonteCarloMetropolis1949} has found an  application in almost every problem involving sampling. 
 Its combination with the theory of Markov chains, the Markov chain Monte Carlo (MCMC) method has lead to an explosion of randomized search algorithms such as \cite{MetropolisAlgorithm, MetropolisHastingsAlgo,GibbsAlgorithm}. 
%
Although  MCMC has been around for decades, algorithm design for fiber exploration is relatively young. 
Standard statistics theory offers asymptotic  solutions to the testing problem, yet they often do not apply in practice. 
 This has been known since the late 1980s \citep{Haberman88}; the fundamental  issue necessitates a synthetic data sample from a fiber associated to the statistical model. Theoretical guarantees for constructing irreducible Markov chains on fibers 
 were provided in the seminal paper \cite{DS98};  while Besag and Clifford argue in  contemporaneous work   \cite{BesagCliffordExchangableSample}  that it is best to produce an exchangeable sample from the fiber  and not rely on MCMC, which may be slow to converge and produce  a good enough sample for testing. 
The last three decades have seen an explosion of research in this area inspired by \cite{DS98}, see \cite{MB25years} for a summary of the state of the art. 
This brings us to our starting point: even with randomization, in applications dealing with high-dimensional polytopes $\mathcal P_M(b)$, there is a computational bottleneck 
 for problems at scale. Unfortunately, this can already happen when  the design matrix $M$ has just hundreds of columns. 

Another line of research that has taken off since the 1950s is reinforcement learning (RL) and optimal control, 
pioneered by Richard Bellman \cite{bellman1954theory}.   Reinforcement learning algorithms can be thought of as informed random walks, in which the agent--or controller--learns from previous experiences through a reward system, in the process refining her understanding of the underlying problem dynamics. Every RL algorithm operates on a Markov decision process (MDP) or partially observed MDP. 
In their simplest form, MDPs are characterized by a set of states, actions, and corresponding rewards. The goal of an agent is to optimize a given objective, also known as the state-value function of the MDP. In this process, the agent learns the optimal policy, or control, which is a function mapping states to actions. 
In general, RL algorithms can be split into actor (policy-based), critic (value-based), and actor-critic (policy-value-based); further, they can be classified as  on-policy or off-policy algorithms. Some of the earliest critic algorithms include policy and value iterations \cite{howard1960dynamic} which use iterative updates to refine the agent's policy and state-value function. Another common family of value-based algorithms  
uses temporal differences (TD), computed by the Bellman equation. A representative algorithm of this family is Q-learning developed by Watkins \cite{WatkinsQL}, which uses an action-value instead of a state-value function; other notable work 
 includes \cite{sutton1988learning} and \cite{tsitsiklis1996analysis}. All of these try to learn a certain score function first, and then subsequently extract the optimal policy.  
In contrast, in  actor algorithms, the agent seeks to directly find the best policy which optimizes the given objective.  Policies are usually parameterized by neural networks and the parameters are updated through gradient descent \cite{Sutton1999PolicyGM}. Other parametrizations can be found in \cite{StochasticApproxMCGlynn}, \cite{RLPOMDPJaakkola}, \cite{SimulationBasedOptimizationsMarbachTsitsiklis}, and \cite{Williams2004SimpleSG}. What makes such algorithms attractive is that they do not suffer from the curse of dimensionality, as neural networks can handle high dimensional data.  
Finally, actor-critic  algorithms \cite{BartoSuttonACfirst}  improve performance by  taking advantage of both appoaches: 
they use   both the parametrization of the policy and temporal difference updates of the state-value function to solve the given MDP. Notable work in this area includes \cite{Konda1999ActorCriticA}, \cite{tsitsiklis1996analysis}.

\subsection{State of the art of fiber sampling}
Reflection on the statistics literature indicates that RL can have an impact on fiber sampling. Namely, the biggest drawback of the  popular algorithm  \cite{DS98} for performing exact tests is that it requires one to compute a \emph{non-linear-algebra object:} a Markov basis of the design matrix $M$ for the polytope $P_M(b)$; see  \cite{WhatIsMB} for a  short introduction. It is worth noting that resorting to subsets of this set is not reliable; e.g., lattice bases (see Definition \ref{defn: lattice basis}) alone are known to be unreliable, see \cite[Example 3.13]{MB25years}. Markov bases can be arbitrarily complicated, as shown in \cite{DeLoeraOnnBadNews}. 
Even worse, computing the full Markov basis for any given large problem is impossible; the only scalable samplers build Markov bases dynamically. Hundreds of papers have been published on Markov bases structure;  \cite{MB25years} is a recent survey. We note the existence of a broader literature on fiber sampling based on, e.g., sequential importance sampling (SIS). \cite{Dob2012} argues that the  Markov bases approach computed dynamically performed better than SIS; see also \cite[Remark 2.2]{RUMBA}.




\section{The Markov Decision Process(MDP)} \label{section:setup}

\subsection*{Fiber basics} 
Fix  the integer design matrix $M\in\mathbb Z^{n\times d}$, 
and one lattice point  $u\in\mathbb Z_{\geq0}^d$. This point $u$ will represent the \emph{observed categorical data}; our statistics readers may think about  a vector flattening of a contingency table, which is a $k$-dimensional tensor of format $d_1\times \dots, \times d_k$ with integer entries, in which case $d:=d_1\cdots d_k$. 
Consider the   set of integer points in the polytope $\mathcal P_M(b)$ such that $b=Mu$. The vector $Mu$ is the observed value of the sufficient statistics for the model, corresponding to the observed data $u$.   

\subsection*{What it takes to move} 
Any two points $v_1,v_2\in\F{M}(Mu)$ satisfy $M(v_1-v_2)=0$. Any vector $v_1-v_2\in\ker M$ is called a \emph{move} on the fiber. The vector space basis of the lattice $\ker M$ is any set of  $n-rank(M)$ vectors in $\ker M$ that are linearly independent; we call any such spanning set \emph{the lattice basis} of $M$. For a summary of the many bases of the integer lattice $\ker M$, the reader is invited to consult \cite[\S 1.3]{DSS09}. We view any lattice basis as a set of connecting moves as follows: 
\begin{definition}[Lattice basis]\label{defn: lattice basis}
    A lattice basis of a design matrix $M$ is any set of $c:=n-rank(M)$ linearly independent moves $\B(M):= \{ l_1,\dots,l_c\} \subset \mathbb{Z}^d$, $Ml_i = 0$.
    For any data  $u\in \mathbb{Z}_{\geq 0}^d$ and for any $v\in \F{M}(Mu)$, there exist $l_{i_1},\dots,l_{i_k}\in \B(M)$ and corresponding scalars $\alpha_{i_1},\dots, \alpha_{i_k} \in \mathbb{Z}$ such that the linear combination can be used to reach $v$ from $u$:   \quad
\(
        u+\alpha_{i_1}l_{i_1}+\dots + \alpha_{i_k}l_{i_k} = v
\). 
\end{definition}
Note that any integral linear combination of elements of $\B(M)$ is also a move: \begin{equation}\label{eqn: lin combo of lattice basis} \alpha_{i_1}l_{i_1}+\dots + \alpha_{i_k}l_{i_k} \in\ker M. \end{equation}
Lattice bases alone \emph{cannot} be used to directly create a random walk on the fiber due to the fact that any sub-sum in the above equation may result in negative entries in the vector.  This brings us to the definition of  Markov bases: 
    any set of moves $\B:= \{ b_1,\dots,b_m\} \subset \mathbb{Z}^{d}$, $m\geq c$, $Mb_i = 0$, such that for any data  $u\in \mathbb{Z}_{\geq 0}^d$ and for any $v\in \F{M}(Mu)$,  there exist $b_{i_1},\dots,b_{i_N}\in \B$ that can  reach $v$ from $u$ \emph{while being on the fiber}: 
\begin{align*}
    &u+b_{i_1}+\dots + b_{i_N} = v, \mbox{ with the condition that } \\
    & u+b_{i_1}+\dots + b_{i_j}\in\mathbb Z^d_{\geq 0} \mbox{ for all }j\leq N.
\end{align*}
The power of Markov bases is they exist and are finite for any integer matrix $M$ by the Hilbert basis theorem, and they guarantee a connected Markov chain on all fibers of $M$ simultaneously. This notion, introduced in  \cite{DS98}, has generated close to a 1000 research papers to date. The following points  are essential for us: a Markov basis is the \emph{minimal} set of moves that guarantees fiber connectivity, but it does so for every possible fiber of a given $M$, i.e., all possible observed data $u$. For a given  $u$, one rarely needs all of the basis elements to connect that specific fiber \cite{Dob2012}.  Markov bases can be arbitrarily complicated, but there are bounds on their complexity in  special cases; see \cite{N-fold2008}, \citep{MB25years}. 
A lattice basis of $M$ is a minimal sub-basis in the sense of  \cite[Lemma 4.2]{CDS-SISmultiwayTables2006}):  $\B(M)$ generates the lattice as a vector space and the saturation of the corresponding ideal results in the full toric ideal; while the detailed algebraic discussions are out of scope for this work, the interested reader can find details in \cite{St} and \cite{DS98}.  A Gr\"obner basis \cite{StWhatIsGB} of  $M$ always contains a minimal Markov basis. 

\subsection*{Recasting fiber moves into an RL framework} 
Any set of moves defined above can be used as a set of actions for exploring the fiber in the context of reinforcement learning. In terms of scalability, lattice bases are the only bases that can be computed for very large matrices $M$, because other sets require non-linear algebra computations.  
With this in mind, we define the MDP of interest.

\begin{definition}(Fiber Sampling Markov Decision Process)\label{defn:MDP}
Fix an integer matrix $M\in\mathbb Z^{n\times d}$ and  $S_0\in\mathbb Z_{\geq 0}^d$, a given initial solution determining the fiber $\fiber{M}(Ms_0)$. 
 A \emph{Fiber Sampling Markov Decision Process} with planing horizon $T=\infty$ is a stationary process  on data $(\St,\St^{-},\St^{0},\At,
 \Qt,R_t,\gamma)$, for $t=0,1,\dots$, defined as follows. 

The state space $\St :=  \fiber{M}(Ms_0)$  consists of all integral points in the polytope $\mathcal P_M(b)$.  
The set of unfeasible states is $\St^{-}: = \{ s\in \mathbb{Z}^d\ |\ Ms = b,  s\not\geq 0\}$, while  
$\St^{0}$ is the set of discovered feasible states. At the start of the algorithm,  $\St^{0} = \{s_0\}$. 

The action space $\At$ consists of \emph{moves} derived from the linear combination of lattice basis vectors, so that each action is given as $a = \alpha_1l_1 + \dots \alpha_c l_c $, with $l_i\in\B(M)$. 
The random variable whose realization at time $t$ is action $a$ is denoted by $A_t$.

The transition kernel is deterministic; i.e., the transition dynamics are simply 
\( 
S_{t+1} = 
 S_t + A_t.
\) 

The mapping 
$R_{t+1}:\ \St\times \At \rightarrow \mathbb{R}$  is the undiscounted reward collected after time $t$, and 
$\gamma$ is the discount factor. 
The reward function $R_{t+1}$, obtained after time $t$ is defined by two sub-rewards. Namely,
\(
        R_{t+1}(S_t,A_t) = R^{feas}_{t+1}(S_t,A_t) + R^{zero}_{t+1}(S_t,A_t)
\)
    where 
\begin{align}\label{eqn: reward split}
        R^{feas}_{t+1}(S_t,A_t) &= \sum_{i = 1}^d I_{ (S_t+A_t)_i < 0}(S_t+A_t)_i\\
        R^{zero}_{t+1}(S_t,A_t) &=
        \begin{cases}
        -d &\text{ if } A_t\equiv \textbf{0}\\
        0 &\text{ otherwise }.
        \end{cases}
\end{align}

The maximization objective is given by 
\begin{equation}\label{eqn: max_objective_IFP}
    \max_{\pi} \EXP_{\pi}\Big[ \sum_{t=0}^\infty \gamma^t R_{t+1}(S_t,\pi(A_t|S_t)) \Big], 
\end{equation}
where $\tau = (s_0,a_0,s_1,a_1,\dots)$ is a trajectory sampled under policy $\pi$. 
     The probability of observing a trajectory $\tau \in \Omega$ is computed as: 
\begin{align*}
    \Qt(\tau) &= \Qt(S_0)\pi(A_0|S_0)\prod_{t=1}^\infty \Qt(S_{t}|S_{t-1},A_{t-1}) \pi(A_t|S
    _t) \\
    &= \Qt(S_0)\pi(A_0|S_0)\prod_{t=1}^\infty \pi(A_t|S
    _t).
\end{align*}

\end{definition}
In practice, in the definition above, because we are treating all future states with almost same importance, we set $\gamma=0.99$. If for whatever reason, the algorithm designer thinks that current states have more meaning than the future states, then $\gamma << 1$ (an example of such a value choice would be if the value of money is lost over long horizons).

We will denote  a truncated finite trajectory by  $\tau_{t:t+h} := (s_t,a_t,\dots,s_{t+h-1},a_{t+h-1},s_{t+h})$. 

In Equation~\eqref{eqn: reward split}, $I$ denotes the indicator function,  so $R^{feas}_{t+1}$  
penalizes the agent based on the magnitude of negative coordinates in $\St^{-}$, if the action vector is not feasible. On the other hand, $R^{zero}_{t+1}$ penalizes the agent for picking the zero action, which would not let   agent explore the environment. 

\label{action_space_page_remark}
A remark about the action space in practice is in order. 
A critical problem  \cite{wu2022turning} faced was the size of the action space. They used a lifting of the Gr\"obner basis of the 3-way transportation polytope specified by the linear embedding, and considered actions with entries in $\{ -1,0,1 \}$. Their agent's policy $\pi$ produces the actions in the form of tables directly from a neural network. 
 In our case, we  use   linear combinations of lattice basis elements to produce  actions. However, such set of moves is also  very large, given the range of  coefficients $\alpha_i$. Namely, the best universal upper bound on the $\alpha_i$ is too large (deriving, e.g., from \cite[Theorem 4.7]{St}). In this case, convergence of the RL algorithm will be extremely slow due to the fact that the agent has to try each action for each state it encounters. Therefore, we  consider moves with coefficients with pre-specified bounds: 
\[
        \forall i\in[N],\ ~\min \alpha_i = c_1,\ \max \alpha_i = c_2,\text{ s.t. } c_1 < c_2\in \mathbb{Z}. 
\]    
Our agent's policy only produces coefficients  $\alpha_i$, which are then used to create a move as a linear combination of lattice bases as in Equation~\eqref{eqn: lin combo of lattice basis}. 

Theoretically, $c_1$ and $c_2$ are known to be finite, resulting in a finite action space, which is needed for the proof of convergence of the actor-critic algorithm. In practice, we may choose the constants $c_1$ and $c_2$ to be smaller than the very large upper bound, to speed up the algorithm. 
As we will see on the examples, the algorithm will work on high dimensional sparse problems, which can be attributed to two things. First, picking coefficient bounds to be at most $2$, and having thousands of lattice basis vectors, is enough to construct moves with both small and large $L_1$ norms. If the initial point is sparse and has small $L_1$ norm, such a coefficient bound is going to help the algorithm find local moves around the initial point and actually move into a different region of the fiber. Secondly, considering the reward design, we observe that no matter the size of the moves we allow the agent to use, the objective of the MDP forces the agent to lower the $L_1$ norm on its moves to the extent where sampling on high dimensional, sparse points is feasible. 
In conclusion, we use low upper bounds on coefficients in order to help the agent pick smaller moves, and in turn learn more efficiently. However, in theory this is not necessary, and given long enough training periods and due to the penalty of the reward, the  agent will slowly learn to use small coefficients in order to successfully sample the fiber.


\section{Actor-critic algorithm for fiber sampling} \label{sec: algorithm} 


\subsection{The policy} 
The policy $\pi$ is typically modeled by a parameterized function $\pi_{\actorparam{}}$. 
In our case, the parameterized policy is constructed out of two fully connected deep neural networks. 
The first part is called the feature extractor, it takes as input the current state $S_t = s$ at time $t$, and it tries to extract certain features shared by all elements in the state space.
The second part is the policy network $h_{\mu, \sigma}(s,a; \actorparam{})$ which outputs the mean and variance which is used to obtain a probability distribution $\pi_{\actorparam{}}(a|s) := \mathcal N(\mu_{\theta}, \sigma_{\theta})$, over the action space. After generation, the outputs are rounded to nearest integers in order to be applicable to the current state. We note that our algorithm learns the approximate optimal stochastic policy which maximizes the objective function. This allows us to use the trained policy as the basis of a sampler which can be retrofitted with additional parts to resemble, for example a Metropolis-Hastings algorithm. 
The part of the architecture in Figure~\ref{fig:Architecture} depicted by full lines refers to the policy, with $\phi_{\actorparam{}}$ representing the feature extractor.  
%
\begin{assumption}\label{assumption: boundedPolicyParam}
The subset $\Theta \subset \mathbb{R}^n$ of network parameters is closed and bounded. 
\end{assumption}
 This is perhaps the strongest assumption we impose throughout this work. It is consistent with  \cite{PPOconvergence}, which assumes compact support of networks over the parameter space. In practice, this is satisfied if one takes a large enough closed ball around initial parameters $\actorparam{0}$.
 
\begin{claim}\label{claim: boundedGrad}
    (a) For every  $s\in \St$ and $a\in \At$, \quad $\pi_{\actorparam{}}(a|s) > 0$. (b) For every  $(s,a) \in \St \times \At$, the mapping $\actorparam{} \mapsto \pi_{\actorparam{}}(a|s)$ is twice continuously differentiable. Furthermore, the $\mathbb{R}^n$-valued function $\actorparam{} \mapsto \nabla_{\actorparam{}}\ln\pi_{\actorparam{}}(a|s)$ is bounded and has a bounded first derivative for any fixed $s$ and $a$. 
\end{claim}
That the gradient is bounded essentially follows from  the policy definition above and finiteness of the state and action spaces. Detailed proof is in Appendix~\ref{appendix:bounded gradient}.


\subsection{The actor} 
The minimization objective of the actor is given by the objective function $L_{a}(\actorparam{})$, defined as $L_{a}(\actorparam{}):= V_{\pi_{\actorparam{}}}(s_0) $, which is exactly the state-value function starting from the initial state $S_0 = s_0$. Using the Policy Gradient Theorem \cite{Sutton2005ReinforcementLA}, we can compute the gradient of the objective function as $ \nabla_{\actorparam{}} L_{a}(\actorparam{}) = \nabla_{\actorparam{}} V_{\pi_{\actorparam{}}}(s_0) = \EXP_{\pi_{\actorparam{}}}\Big[ Q_{\pi_{\actorparam{}}}(S_t,A_t) \nabla_{\actorparam{}}\ln \pi_{\actorparam{}}(A_t|S_t) \Big]$.
We can modify the gradient computation above in a way that the value of the gradient remains the same, while the variance of an estimator of this gradient is reduced. We accomplish this by introducing a baseline function, denoted $b(\cdot)$, which doesn't depend on the action. The gradient is then $\nabla_{\actorparam{}} L_{a}(\actorparam{}) = \nabla_{\actorparam{}} V_{\pi_{\actorparam{}}}(s_0) = \EXP_{\pi_{\actorparam{}}}\Big[ (Q_{\pi_{\actorparam{}}}(S_t,A_t)-b(S_t))\nabla_{\actorparam{}}\ln \pi_{\actorparam{}}(A_t|S_t) \Big]$. A popular choice is $b(S_t) = V_{\pi_{\actorparam{}}}(S_t)$, which gives us the the definition of the advantage function $A_{\actorparam{}}(S_t,A_t) := Q_{\pi_{\actorparam{}}}(S_t,A_t) - V_{\pi_{\actorparam{}}}(S_t)$. This function measures  how good or bad the action $A_t$ is at  state $S_t$, compared to the policy $\pi_{\actorparam{}}$. We take the derivation one step further and introduce the Generalized Advantage Estimator $\widehat{A_{GAE}}(t)$ \cite{schulman2015high}; as the name suggests, it estimates the advantage function. Using $\widehat{A_{GAE}}(t)$ with Fubini-Tonelli theorem, in place of $A_{\actorparam{}}(S_t,A_t)$, we define the following policy gradient: 
\begin{align*}
    \nabla_{\actorparam{}} L_{GAE}(\actorparam{}) &:= \EXP_{\pi_{\actorparam{}}}\Big[\widehat{A_{GAE}}(t)\nabla_{\actorparam{}}\ln\pi_{\actorparam{}}(A_t|S_t)\Big] \\
    &= \sum_{l=0}^\infty (\gamma \lambda)^l \EXP_{\pi_{\actorparam{}}}\Big[\delta(t+l)\nabla_{\actorparam{}}\ln\pi_{\actorparam{}}(A_t|S_t)\Big].
\end{align*}

All the above computations have been derived under the assumption that we have an exact expression of the state-value function $V_{\pi_{\actorparam{}}}$. However, in practice, this is usually not the case and we can only work with an approximation of such a function under some parametrization $\criticparam{}$. In what follows, we show that such a parametrization does not pose any issue in the gradient estimate. We follow the work of \cite{actorCriticAlgos} and \cite{tsitsiklis1996analysis} and define, for any $\actorparam{}$, two inner products $\langle \cdot, \cdot \rangle_{\theta}^1$ and $\langle \cdot, \cdot \rangle_{\theta}^2$ of two pairs of real-valued functions $F_1^1,F_2^1$, on $\mathcal{S}\times \mathcal{A}$, and $F_1^2,F_2^2$ on $\mathcal{S}$ viewed as vectors in $\mathbb{Z}^{|\mathcal{S}||\mathcal{A}|}$ and $\mathbb{Z}^{|\mathcal{S}|}$, by 
\begin{align*}
    &\langle F_1^1, F_2^1 \rangle_{\theta}^1 := \sum_{s,a}\mu_{\theta{}}(s)\pi_{\theta}(a|s) F_1^1(s,a)F_2^1(s,a)\\
    &\langle F_1^2, F_2^2 \rangle_{\theta}^2 := \sum_{s}\mu_{\theta{}}(s)F_1^2(s)F_2^2(s).
\end{align*}
Define projections $\Pi^1_{\actorparam{}}: \mathbb{Z}^{|\St||\At|} \mapsto \Psi_{\actorparam{}}^1$, $\Pi^2_{\actorparam{}}: \mathbb{Z}^{|\St|} \mapsto \Psi_{\actorparam{}}^2$ as $\Pi^1_{\actorparam{}} F := \arg\min_{\hat{F}\in \Psi_{\actorparam{}}^1}\norm{F - \hat{F}}{\actorparam{}}^1$ and $\Pi^2_{\actorparam{}} F := \arg\min_{\hat{F}\in \Psi_{\actorparam{}}^2}\norm{F - \hat{F}}{\actorparam{}}^2$, respectively. 
Here, $\norm{\cdot}{\actorparam{}}^1$ and $\norm{\cdot}{\actorparam{}}^2$ are the norms on $\mathbb{Z}^{|\St||\At|}$,$\mathbb{Z}^{|\St|}$ induced by inner products $\innerProd{\cdot}{\cdot}{\actorparam{}}^1$ and $\innerProd{\cdot}{\cdot}{\actorparam{}}^2$. By definition of a projection, it follows that the error of the projection is orthogonal to the projection space. Hence, in both cases, $\innerProd{\Pi^1_{\actorparam{}}R_{t+1}(S_t,A_t) - R_{t+1}(S_t,A_t) }{\psi_{\actorparam{}}^{1,i}}{\actorparam{}}^1 = 0$ and   $\innerProd{\Pi^2_{\actorparam{}}V_{\pi_{\actorparam{}}} - V_{\pi_{\actorparam{}}} }{\psi_{\actorparam{}}^{2,i}}{\actorparam{}}^2 = 0$ imply that $\innerProd{\Pi^1_{\actorparam{}}R_{t+1}(S_t,A_t)}{\psi_{\actorparam{}}^{1,i}}{\actorparam{}}^1 = \innerProd{R_{t+1}(S_t,A_t)}{\psi_{\actorparam{}}^{1,i}}{\actorparam{}}^1$ and $\innerProd{\Pi^2_{\actorparam{}}V_{\pi_{\actorparam{}}}}{\psi_{\actorparam{}}^{2,i}}{\actorparam{}}^2 = \innerProd{V_{\pi_{\actorparam{}}}}{\psi_{\actorparam{}}^{2,i}}{\actorparam{}}^2$
which in turn shows that it suffices to learn the projection of $V_{\pi_{\actorparam{}}}$ onto $\Psi_{\actorparam{}}$ in order to compute the approximation of the gradient $\actorgrad(\actorparam{})$. Let $V_{\criticparam{}}$ be a projection of the state-value and let $\delta_{\criticparam{}}(t)$ denote the $t$-th temporal difference which uses the projected state-value function. Then we write the gradient approximation of the exact gradient as $ \actorgrad(\actorparam{},\criticparam{}) := \sum_{l=0}^{K-1}(\gamma \lambda)^l \EXP_{\pi_{\actorparam{}}}\Big[\delta_{\criticparam{}}(t+l)\nabla_{\actorparam{}}\ln\pi_{\actorparam{}}(A_t|S_t)\Big]$
which is estimated as $    \actorgradestim(\actorparam{},\criticparam{}) := \frac{1}{K}\sum_{k=0}^{K-1}\nabla_{\actorparam{}}\ln\pi_{\actorparam{}}(A_k|S_k)\sum_{l=k}^{K-1} (\gamma \lambda)^{l-k}\delta_{\criticparam{}}(t+l)$.
Finally, the actor's parameter update is given by the following. 
\begin{definition}[Actor update]\label{def: actorUpdate}
Let $\actorparam{t}$ denote the policy parameter (actor) at time $t$. Then, we define the actor update along a truncated trajectory $\tau_{t:t+K}$ as
\begin{align}
    &\actorparam{t+K} = \actorparam{t+K-1} + \alpha(t+K-1) \nonumber \\
    &\cdot \Big[ \frac{1}{K}\sum_{k=0}^{K-1}\nabla_{\actorparam{}}\ln\pi_{\actorparam{}}(A_{t+k}|S_{t+k})\sum_{l=k}^{K-1} (\gamma \lambda)^{l-k}\delta_{\criticparam{}}(t+l)\Big]. \label{eqn: a2cScheme1}
\end{align}    
\end{definition}

The mathematical details of the derivation in this section can be found in Appendix~\ref{appendix: actor parameter update}. 
\subsection{The critic}
\label{section: critic}

Not having access to the analytical form of  state-value function $V$, we parametrize it: 
\[
    V_{\criticparam{}}(S_t) := \phi_{\actorparam{}}(S_t)^T\criticparam{}\approx V_{\actorparam{}}.
\]
\begin{assumption}\label{assumption: boundedCriticParam}
	The subset $\Omega \subset \mathbb{R}^m$ of critic parameters is closed and bounded.
\end{assumption}
This assumption, as  \ref{assumption: boundedPolicyParam}, is consistent with assumptions in \cite{PPOconvergence}. In practice, this is satisfied by taking a large enough closed ball around initial parameters $\criticparam{0}$.
 
The feature vector $\phi_{\actorparam{}}$ depends on the actor parameter $\actorparam{}$ and is shared between the actor and the critic. Such setup is a deviation from \cite{tsitsiklis1996analysis} and most of the other actor-critic algorithms. To the best of our knowledge, the convergence of algorithms that use this architecture has only been explored in \cite{actorCriticAlgos}. Pictorially, the parameterization of the critic is given by the dashed pipeline in Figure \ref{fig:Architecture}, from the state  $\St$, through the feature extractor parametrized by $\phi_{\actorparam{}}$,  and finally to the critic network $\criticparam{}^T\phi_{\actorparam{}}$. 
%
The activation functions of the feature extractor are chosen such that the mapping $\actorparam{} \mapsto \phi_{\actorparam{}}$ is continuous, bounded and differentiable. The full claim with proof can be found in Appendix \ref{appendix: feature vector}.

To enable the critic to learn the approximate state-value function parametrized by $\criticparam{}$, we need to define a correct minimization objective for the critic.  
In our setup, the algorithm is learning the necessary weights in an on-line manner, meaning that the only observations that the algorithm can use to learn are the ones that are observed along some sampled trajectory. Work in \cite{Sutton2005ReinforcementLA} gives us several choices of minimization objectives which work with incremental data. 
One of the most basic but well established minimization objectives is the $n-$step bootstrap, which minimizes the difference between collected rewards along a truncated trajectory $\tau_{t:t+K}$ and the parametrized state-value function $V_{\criticparam{}}$.
The detailed formulas appear in Appendix~\ref{appendix: critic}. This leads to the following: 
\begin{definition}[Critic Minimization Objective]
The minimization objective for the critic is: 
\begin{align}\label{eqn: critic minimization objective}
    &L_{\criticgrad}(\actorparam{},\criticparam{}) =\nonumber \\
    &\min_{\criticparam{}} \frac{1}{2}\EXP_{\pi_{\actorparam{}}}\Big[\sum_{k=0}^{K-1}\Big( - V_{\criticparam{}}(S_{t+k})\nonumber\\
    &+ \big(\sum_{j=k}^{K-1} \gamma^{j-k} R_{t+1+j}(S_{t+j},A_{t+j})  +\gamma^{K-k} V_{\criticparam{}}(S_{t+K})\big) 
 \Big)^2\Big].
\end{align}
\end{definition}

\begin{definition}[Critic update]\label{def: criticUpdate}
    Let $\criticparam{t}$ denote the state-value parameter (critic) at time $t$. Then, we define the critic update along a truncated trajectory $\tau_{t:t+K}$ as
    \begin{align}
    &\criticparam{t+K} = \criticparam{t+K-1}+ \beta(t+K-1) \nonumber \\
    &\cdot \sum_{k=0}^{K-1}\Big( V_{\criticparam{t+K-1}}(S_{t+k})- \big(\sum_{j=k}^{K-1} \gamma^{j-k} R_{t+1+j}(S_{t+j},A_{t+j})  \nonumber\\
    &+\gamma^{K-k} V_{\criticparam{t-K+1}}(S_{t+K})\big)\Big)\Big(\phi_{\actorparam{t+K-1}}(S_{t+k}) - \gamma^{K-k}\phi_{\actorparam{t+K-1}}(S_{t+K}) \Big)\label{eqn: a2cScheme2},\\
    & \nabla_{\criticparam{t+K-1}}V_{\criticparam{t+K-1}}(S_{t+l}) = \phi_{\actorparam{t+K-1}}(S_{t+l})\nonumber.
    \end{align}    
\end{definition}

The intuition behind the critic parameter update is explained in Appendix~\ref{appendix: critic}.

\subsection{The actor-critic algorithm} 

We combine parameter updates in Definitions~\ref{def: actorUpdate} and~\ref{def: criticUpdate} under the architecture illustrated in Figure~\ref{fig:Architecture}. The full arrows depict the flow of state observations from the environment through the feature extractor and  the actor's network. The dashed arrows depict the flow of state observations from the environment through the feature extractor and  the critic's network. The gradient computation in \ref{def: actorUpdate} and \ref{def: criticUpdate} is executed independently with addition that the actor's back-propagation is used to update the weights in the feature extractor as well. 
Pseudo code can be found in Appendix~\ref{appendix: pseudocode}. 

\begin{figure}[H]
    \centering
    \begin{tikzpicture}
\draw[thick,->] (-2.5,0.1) -- (-1.5,0.1); 
\draw[dashed,thick,->] (-2.5,-0.1) -- (-1.5,-0.1); 
    \draw (-1,-0.5) -- (0,-0.5) -- (0,0.5) -- (-1,0.5) -- (-1,-0.5); 
    \draw[thick,->] (0.5,0) -- (1,0.7); 
    \draw (1,0.3) -- (2,0.3) -- (1.5,1.3) -- (1,0.3); 
    \draw[thick,->] (2.2,0.7) -- (3.2,0.7); 
    \draw[dashed,thick,->] (0.5,0) -- (1,-0.7); 
    \draw (1,-1) -- (2,-1) -- (1.5,0) -- (1,-1); 
    \draw[dashed,thick,->] (2.2,-0.7) -- (3.2,-0.7); 

    \node at (4,-0.7) {$V_{\criticparam{}}(s)$};
    \node at (4, 0.7) {$\pi_{\actorparam{}}(a|s)$};
    \node at (-3,0) {$\St$};
    \node at (1.5,0.5) {$\pi_{\actorparam{}}$};
    \node at (1.5,-0.8) {$\criticparam{}^T\phi_{\actorparam{}}$};
    \node at (-0.5,0) {$\phi_{\actorparam{}}$};
\end{tikzpicture}
    \caption{Actor-Critic Architecture: Actor and critic share features $\phi_{\actorparam{}}$ and use them separately to produce the current action and state-value, respectively.}
    \label{fig:Architecture}
\end{figure}

The code implementing our actor-critic algorithm is available on the following link:

\url{https://github.com/ivangvozdanovic/Fiber-Sampling-Using-Reinforcement-Learning/tree/main}

\section{Convergence analysis} 
 \label{section: convergence_proof}

For a given matrix $M\in\mathbb Z^{n\times d}$ and a starting point $u\in\mathbb Z_{\geq0}^d$, we will refer to the algorithm which computes a lattice basis of $M$ and applies A2C via architecture~\ref{fig:Architecture} as the \emph{``Actor-Critic Sampler"} (Algorithm~\ref{alg:A2CMoveGen}). 

\begin{theorem} \label{thm: main}
    Assume that the parameter spaces $\Theta$ and $\Omega$ are closed and bounded. Given the finite fiber $\fiber{M}$, bounded reward $R_t$, and parameter updates in Equations~\eqref{def: actorUpdate} and~\eqref{def: criticUpdate}, the Actor-Critic Sampler  converges to the approximate optimal policy $\pi^*$ for exploring the fiber using the MDP~\ref{defn:MDP}: 
    \begin{align*}
        \pi^* = \arg\max_{\pi}\EXP_{\pi}\Big[ \sum_{t=0}^\infty \gamma^t R_{t+1}(S_t,\pi(A_t|S_t)) \Big].
    \end{align*}
\end{theorem}

\paragraph{Proof intuition.} 
The proof of convergence of the Actor-Critic Sampler algorithm relies on the 2-timescale ordinary differential equation analysis pioneered by \cite{BORKAR1997291}. This method has been used to analyze various versions of the actor-critic method; see  \cite{BorkarStochasticApprox},\cite{ramaswamy2018stability}, \cite{borkar2009stochastic}. 
 The tools that we use to show convergence rely on 6 technical conditions, all of which are  
stated and proved in Appendix \ref{appendix: convergence_proof}, pages~\pageref{first page of proof}-\pageref{last page of proof}. 
 Given two assumptions (\ref{assumption: boundedPolicyParam}, and \ref{assumption: boundedCriticParam}) on the compactness of the parameter space of the actor and critic, as well as a standard assumption on the ergodicity (\ref{assumption: ergodicity}) of the Markov chains simulated in the Markov decision process from Definition~\ref{defn:MDP}, we show that our algorithm satisfies these 6 conditions and hence converges.

To address the  computational complexity of  determining a lattice basis for  large  $M$,   note the following.  
\begin{corollary}\label{cor: reconstruction} 
Optimality is preserved under subdividing the problem, applying the  Actor-Critic Sampler, and embedding the resulting policy into the original problem. 
\end{corollary}
Exactly how problem subdivision is achieved may    depend on the  particular problem and data structure.  Instances need to be small enough so that lattice bases can be computed,  but also contain enough information to yield a conclusive answer to the proposed statistical test.  For network data, we tested this idea using three different methods for initial data subdivision: $k$-core decomposition, bridge cuts, and induced subgraphs. 

\section{Experiments}\label{sec: simulations} 
Data generation and computing times are detailed in Appendix~\ref{appendix:experimentData}, followed by pseudocode for the Actor-Critic Sampler.

\subsection*{Performance evaluation metrics}  
With the statistical testing goal in mind, we report goodness-of-fit test conclusions for exact conditional  test  and compare it with known  results from the literature. 
When results are not known and we are exploring the scalability of the  Actor-Critic Sampler, we report two other  metrics:  the number of discovered states in the fiber given different levels of sparsity,  which demonstrates  we can obtain an exchangeable fiber sample and scale the procedure; and    the number of discovered states in the fiber which the agent hasn't learned on,  looking to deduce how well the algorithm generalizes to unseen data. 
We report computation times along with each example. 

\begin{figure}[!ht] 
\centering
 \includegraphics[width=0.45\linewidth]{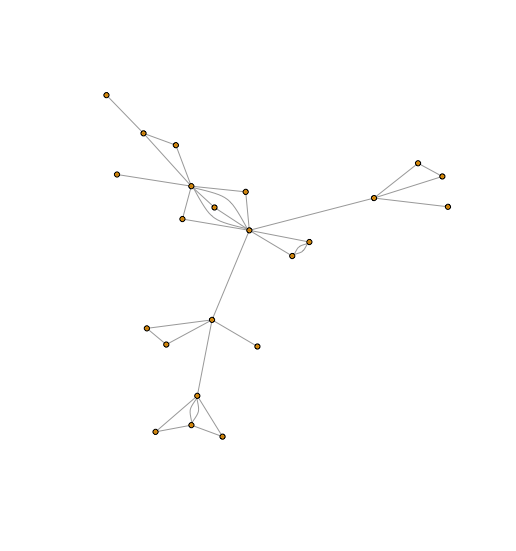}   \includegraphics[width=0.5\linewidth]{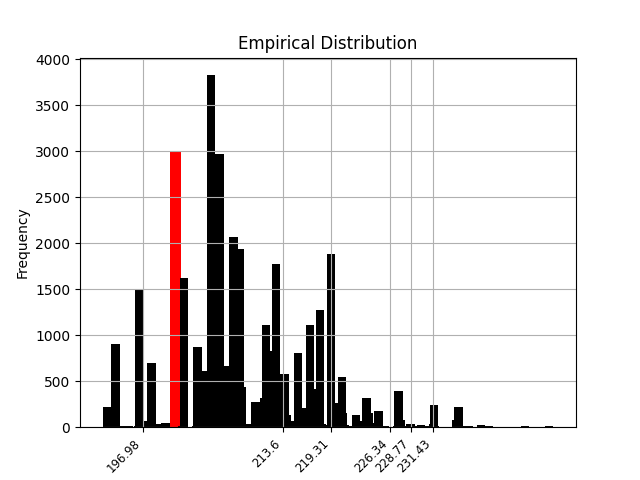} 
\caption{(Left) A component of  statistician coauthorship network with $23$ nodes.  (Right) Sampling distribution of the chi-square statistic on the fiber; sample size $50,000$. The observed value is in red.  
}
\label{fig: chi_square_dist_p_val} 
\end{figure}

\subsection*{Results} We evaluate  algorithm performance, scaling, and generalization;  following are four main takeaways. All  $p$-values are computed referencing the sampling distribution of the $\chi^2$ statistic on the fiber. 


{\bf (1)} Consider a data set of applied interest in network analysis: a coauthorship network of statisticians presented in \cite{JiJinAuthorsAOAS}. A model-based approach for analyzing (the large connected component of) this  network, which we use as a benchmark, is offered in  \cite{KarwaPetrovic:AOASauthors}. Data is represented as an adjacency matrix of an undirected graph on $N$ nodes: as  a vertex-edge incidence vector in which $d$ is the number of node-pairs. 
 The model of interest here is the $\beta$-model for random graphs   \cite{chatterjee2011random}, with the design matrix $M\in\mathbb Z^{N\times d}$. 
Our statistical conclusions for model/data fit are  consistent with the known results: the $\beta$ model fits the data well, with the $p$-value 0.9957. The benchmark value from the literature \cite{KarwaPetrovic:AOASauthors} is 0.997.  The sampling distribution of the goodness-of-fit statistic  in Figure \ref{fig: chi_square_dist_p_val} also matches  the benchmark. 

\begin{figure}[H]
    \centering
    \begin{tabular}{cc}
        \subf{\includegraphics[width=65mm]{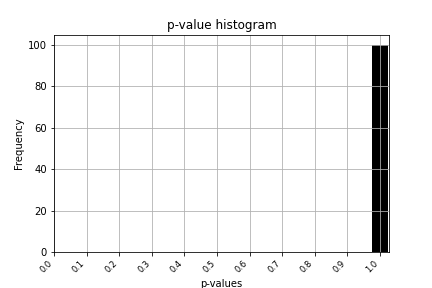}}
         {}
         \subf{\includegraphics[width=65mm]{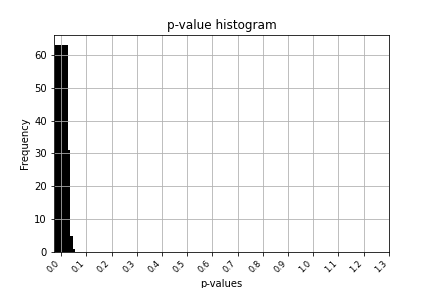}}
        {}
    \end{tabular}
    \caption{Histograms for 100 p-values from 100 exchangeable samples of size 100, obtained through simulation of 100 independent Markov chains using the trained optimal policy $\pi^*$. 
    (Left) The independence model fits  the data from \cite[p.371]{DS98}. 
    (Right) The model does not fit the  data from \cite[Ex.3.13]{MB25years}. Data tables   provided in  appendix.
    }  
    \label{fig: p_value_test}
\end{figure}

{\bf (2)} Two carefully designed synthetic datasets taken from \cite{MB25years}, which show that sampling with lattice basis can fail even in low-dimensional problems, demonstrate the ability of our algorithm to correctly test for model fit on sparse data for which lattice bases samplers have been shown to fail.  We   produce exchangeable samples  and obtain results consistent with  theory; see Figure \ref{fig: p_value_test}.
    Both  computations match the Monte Carlo exact $p$-value estimates and have been verified by asymptotic $p$-value estimates. We compare  Figure~\ref{fig: p_value_test} (Right) in particular  to  two histograms in \cite[Fig.2]{MB25years}, which demonstrate that unlike with Markov bases,  Metropolis-Hastings Markov chains built using lattices bases gives unreliable results. Our algorithm is robust, standing in  stark contrast to the naive lattice-basis sampler. 
 The training time of the policy, using 1000 episodes and  100 steps per episode took 7 minutes for the $0/1$ initial solution and 14 minutes for the initial solution containing larger values. The sampling of Besag-Clifford Markov chains and derivation of the p-value histograms took 34 minutes and 48 minutes, respectively.

\begin{table}[!ht]
 \centering
 \small
\begin{tabular}{ |p{1cm}||p{1.1cm}|p{1.1cm}|p{1.1cm}|p{1.1cm}|||  }
 \hline
 \multicolumn{4}{|c|}{Fiber Exploration} \\
 \hline 
 \diagbox{N}{$p$} & $p = 0.5$ & $p=0.3$ & $p=0.1$ \\
 \hline
 1000 & $6.9e{29}$ & $1.8e{12}$ & $3.6e{9}$ \\ 
 \hline
 2000 &  $2.3e{29}$ & $1.1e{11}$ & $3.7e{7}$ \\ 
 \hline
 3000 & $1.9e{29}$ & $3.4e{29}$ & $4.8e16$ \\ 
\hline
\end{tabular}
\normalsize
 \caption{Illustration of Actor-Critic Sampler scaling, and the negligible effect of data sparsity on fiber exploration.
Table entries are the \emph{number of discovered fiber points} given a  random initial  vector drawn from the  Erd\"{o}s-R\'enyi random graph model $G(N,p)$.}
 \label{tab:fiber_exploration}
\end{table}

\begin{figure}[H]
    \centering
    \begin{tabular}{c}
        \subf{\includegraphics[width=70mm]{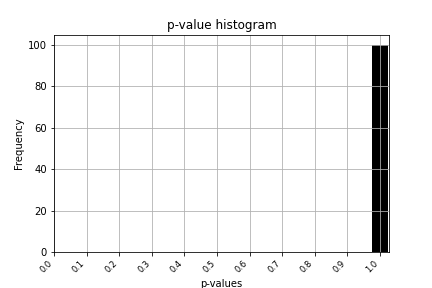}}
         {}
    \end{tabular}
    \caption{Histogram for 100 $p$-values obtained from 100 exchangeable samples of size 100, obtained through simulation of 100 independent Markov chains using the trained optimal policy $\pi^*$.}  
    \label{fig: dobra_p_value}
\end{figure}

{\bf (3)} Structural zeros in contingency tables are known to pose a challenge for fiber sampling and goodness-of-fit testing in general; see \cite{CategoricalDataAnalysis} and  \cite[Proposition 2.1]{HT:10}.     In particular, the usual Markov chain samplers may become disconnected. To this end, \cite[Section 7.1]{Dob2012} offers a detailed study of a $4\times 5\times 4$  cross-classification of $4345$ individuals by occupational groups collected in the 1969 survey of the National Bureau of Economic Research (NBER) from  \cite[p.45]{CategoricalDataAnalysis}.  Using an exchangeable sample from the fiber of the all-two-way interactions model for three categorical random variables, we have reproduced Dobra's conclusion of this goodness-of-fit test; see Figure \ref{fig: dobra_p_value}.  
Our approach was able to train the policy to sample points which obey the structural zero constraints and obtain an exchangeable sample with which we concluded the validity of the proposed underlying model. Optimal policy training for 1000 episodes and 100 steps per episode took 114 minutes while the sampling and p-value histogram computation took roughly 80 minutes.

\vspace{-5mm}

{\bf (4)} To demonstrate scalability of our approach, we simulate data of increasing size and different sparsity levels. 
Table \ref{tab:fiber_exploration} reports the number of discovered fiber points using the Actor-Critic Sampler for subproblems after applying Corollary~\ref{cor: reconstruction}. The algorithm is able to sample a very large number of fiber points under the $\beta$-model of random graphs (in which the sufficient statistic is the graphs's degree sequence), for  Erd\"{o}s-R\'enyi random graphs with  edge probability parameter $p$ and number of nodes $N$. For the samples in Table \ref{tab:fiber_exploration}, we have the following computation times: (N=1000, p=0.5) $\sim$ 10min,  (N=2000, p=0.5) $\sim$ 12.3min,  (N=3000, p=0.5) $\sim$ 24.5min, (N=1000, p=0.3) $\sim$ 25.8min, (N=2000, p=0.3) $\sim$ 41.15min, (N=3000, p=0.3) $\sim$ 65.61, (N=1000, p=0.1) $\sim$ 80.2min, (N=2000, p=0.1) $\sim$ 240.0min, (N=3000, p=0.1) $\sim$ 305.5min.

We close with a note on the large but finite action spaces we work with. Namely, in simulations, we specify constants $c_1$ and $c_2$ (discussed at the end of \S2) in an ad-hoc manner. The literature on fiber sampling for applied statistics supports using \emph{small} values for these constants. For example, small constants are enough to discover all points on fibers $\mathcal F_M(u)$ when $Mu>0$ on graphical models; see \cite{Kahle2012POSITIVEMA}. We continue to discover millions of points in fibers with fairly  values of these constants, and illustrate negligible effect on learning rates.

On a final note, a good set of experiments should seek to find out where a method works and also where it fails.  Thus we remark on the issues we encountered while developing our computational strategy in Appendix~\ref{appendix:exploring failure}.

\bibliographystyle{plain} 
\bibliography{references,AlgStatAndNtwksAndMB,RLforAlgAndOptim}

\begin{thebibliography}{60}
\providecommand{\natexlab}[1]{#1}
\providecommand{\url}[1]{\texttt{#1}}
\expandafter\ifx\csname urlstyle\endcsname\relax
  \providecommand{\doi}[1]{doi: #1}\else
  \providecommand{\doi}{doi: \begingroup \urlstyle{rm}\Url}\fi

\bibitem[Absil and Kurdyka(2006)]{AbsilKurdykaGradientSystm}
P.-A. Absil and K.~Kurdyka.
\newblock On the stable equilibrium points of gradient systems.
\newblock \emph{Systems \& Control Letters}, 55\penalty0 (7):\penalty0
  573--577, 2006.
\newblock ISSN 0167-6911.
\newblock \doi{https://doi.org/10.1016/j.sysconle.2006.01.002}

\bibitem[Almendra-Hern\'andez et~al.(2023)Almendra-Hern\'andez, De~Loera, and
  Petrovi\'c]{MB25years}
F\'elix Almendra-Hern\'andez, Jes\'us~A. De~Loera, and Sonja Petrovi\'c.
\newblock Markov bases: a 25 year update.
\newblock \emph{Journal of the American Statistical Association}, 119\penalty0
  (546):\penalty0 1671--1686, 2023.

\bibitem[Baird(1993)]{baird1993advantage}
Leemon~C Baird.
\newblock Advantage updating.
\newblock Technical report, Technical report wl-tr-93-1146, Wright Patterson
  AFB OH, 1993.

\bibitem[Bakenhus and Petrovi\'c(2024)]{RUMBA}
Miles Bakenhus and Sonja Petrovi\'c.
\newblock Sampling lattice points in a polytope: a {B}ayesian biased algorithm
  with random updates.
\newblock \emph{Algebraic Statistics}, 15\penalty0 (1):\penalty0 61--83, 2024.

\bibitem[Barndorff-Nielsen(1978)]{Barndorff-Nielsen}
O.~Barndorff-Nielsen.
\newblock \emph{Information and Exponential Families: In Statistical Theory},
  volume Reprinted in 2014.
\newblock Wiley Series in Probability and Statistics, 1978.

\bibitem[Barto et~al.(1983)Barto, Sutton, and Anderson]{BartoSuttonACfirst}
Andrew~G. Barto, Richard~S. Sutton, and Charles~W. Anderson.
\newblock Neuronlike adaptive elements that can solve difficult learning
  control problems.
\newblock \emph{IEEE Transactions on Systems, Man, and Cybernetics},
  SMC-13\penalty0 (5):\penalty0 834--846, 1983.
\newblock \doi{10.1109/TSMC.1983.6313077}.

\bibitem[Bellman(1954)]{bellman1954theory}
Richard Bellman.
\newblock The theory of dynamic programming.
\newblock \emph{Bulletin of the American Mathematical Society}, 60\penalty0
  (6):\penalty0 503--515, 1954.

\bibitem[Bengio et~al.(2021)Bengio, Lodi, and
  Prouvost]{MLforCombinOptim-Bengio2021}
Yoshua Bengio, Andrea Lodi, and Antoine Prouvost.
\newblock Machine learning for combinatorial optimization: A methodological
  tour d'horizon.
\newblock \emph{European Journal of Operational Research}, 290\penalty0
  (2):\penalty0 405--421, 2021.
\newblock ISSN 0377-2217.
\newblock \doi{https://doi.org/10.1016/j.ejor.2020.07.063}

\bibitem[Besag and Mondal(2013)]{BesagMondal}
J.~Besag and D.~Mondal.
\newblock Exact goodness-of-fit tests for {M}arkov chains.
\newblock \emph{Biometrics}, 69\penalty0 (2), 2013.

\bibitem[Besag and Clifford(1989)]{BesagCliffordExchangableSample}
Julian Besag and Peter Clifford.
\newblock Generalized {M}onte {C}arlo significance tests.
\newblock \emph{Biometrika}, 76\penalty0 (4):\penalty0 633--642, 1989.
\newblock ISSN 00063444

\bibitem[Borkar(2006)]{BorkarStochasticApprox}
Vivek Borkar.
\newblock Stochastic approximation with controlled {M}arkov noise.
\newblock \emph{Systems \& Control Letters}, 55:\penalty0 139--145, 02 2006.
\newblock \doi{10.1016/j.sysconle.2005.06.005}.

\bibitem[Borkar(1997)]{BORKAR1997291}
Vivek~S. Borkar.
\newblock Stochastic approximation with two time scales.
\newblock \emph{Systems \& Control Letters}, 29\penalty0 (5):\penalty0
  291--294, 1997.
\newblock ISSN 0167-6911.
\newblock \doi{https://doi.org/10.1016/S0167-6911(97)90015-3}

\bibitem[Borkar(2009)]{borkar2009stochastic}
Vivek~S Borkar.
\newblock \emph{Stochastic approximation: a dynamical systems viewpoint},
  volume~48.
\newblock Springer, 2009.

\bibitem[Chatterjee et~al.(2011)Chatterjee, Diaconis, and
  Sly]{chatterjee2011random}
S.~Chatterjee, P.~Diaconis, and A.~Sly.
\newblock Random graphs with a given degree sequence.
\newblock \emph{The Annals of Applied Probability}, 21\penalty0 (4):\penalty0
  1400--1435, 2011.

\bibitem[Chen et~al.(2006)Chen, Dinwoodie, and
  Sullivant]{CDS-SISmultiwayTables2006}
Yuguo Chen, Ian~H. Dinwoodie, and Seth Sullivant.
\newblock Sequential importance sampling for multiway tables.
\newblock \emph{Quality Engineering}, 52:\penalty0 277--278, 2006.

\bibitem[De~Loera and Onn(2006)]{DeLoeraOnnBadNews}
Jes\'us~A. De~Loera and Shmuel Onn.
\newblock Markov bases of three-way tables are arbitrarily complicated.
\newblock \emph{Journal of Symbolic Computation}, 41:\penalty0 173--181, 2006.

\bibitem[{De Loera} et~al.(2008){De Loera}, Hemmecke, Onn, and
  Weismantel]{N-fold2008}
Jes{\'u}s~A. {De Loera}, Raymond Hemmecke, Shmuel Onn, and Robert Weismantel.
\newblock N-fold integer programming.
\newblock \emph{Discrete Optimization}, 5\penalty0 (2):\penalty0 231--241,
  2008.

\bibitem[Diaconis and Efron(1985)]{DiaconisEfron}
Persi Diaconis and Bradley Efron.
\newblock Testing for independence in a two-way table: New interpretations of
  the {C}hi-square statistic.
\newblock \emph{Annals of Statistics}, 13\penalty0 (3):\penalty0 845--874,
  1985.

\bibitem[Diaconis and Sturmfels(1998)]{DS98}
Persi Diaconis and Bernd Sturmfels.
\newblock Algebraic algorithms for sampling from conditional distributions.
\newblock \emph{Annals of Statistics}, 26\penalty0 (1):\penalty0 363--397,
  1998.

\bibitem[Dobra(2012)]{Dob2012}
Adrian Dobra.
\newblock Dynamic {M}arkov bases.
\newblock \emph{Journal of Computational and Graphical Statistics}, pages
  496--517, 2012.

\bibitem[Drton et~al.(2009)Drton, Sturmfels, and Sullivant]{DSS09}
Mathias Drton, Bernd Sturmfels, and Seth Sullivant.
\newblock \emph{Lectures on Algebraic Statistics}, volume~39 of
  \emph{Oberwolfach Seminars}.
\newblock Birkh{\"{a}}user, 2009.

\bibitem[Fienberg(1980)]{CategoricalDataAnalysis}
Stephen~E. Fienberg.
\newblock \emph{The Analysis of Cross-Classified Categorical Data}.
\newblock Springer, Reprinted 2007, 1980.

\bibitem[Fienberg and Wasserman(1981)]{FienbergWasserman1981categorical}
Stephen~E. Fienberg and Stanley~S. Wasserman.
\newblock Categorical data analysis of single sociometric relations.
\newblock \emph{Sociological methodology}, 12:\penalty0 156--192, 1981.

\bibitem[Geman and Geman(1988)]{GibbsAlgorithm}
Stuart Geman and Donald Geman.
\newblock {(1984) Stuart Geman and Donald Geman, "Stochastic relaxation, Gibbs
  distributions, and the Bayesian restoration of images," IEEE Transactions on
  Pattern Analysis and Machine Intelligence PAMI-6: 721-741}.
\newblock In \emph{{Neurocomputing, Volume 1: Foundations of Research}}. The
  MIT Press, 04 1988.
\newblock ISBN 9780262267137.
\newblock \doi{10.7551/mitpress/4943.003.0038}

\bibitem[Glynn(1986)]{StochasticApproxMCGlynn}
Peter~W. Glynn.
\newblock Stochastic approximation for {M}onte {C}arlo optimization.
\newblock In \emph{Proceedings of the 18th Conference on Winter Simulation},
  WSC '86, pages 356--365, New York, NY, USA, 1986. Association for Computing
  Machinery.
\newblock ISBN 0911801111

\bibitem[Goldenberg et~al.(2010)Goldenberg, Zheng, Fienberg, and
  Airoldi]{goldenberg2010survey}
Anna Goldenberg, Alice~X Zheng, Stephen~E Fienberg, and Edoardo~M Airoldi.
\newblock A survey of statistical network models.
\newblock \emph{Foundations and Trends{\textregistered} in Machine Learning},
  2\penalty0 (2):\penalty0 129--233, 2010.

\bibitem[Haberman(1988)]{Haberman88}
Shelby~J. Haberman.
\newblock A warning on the use of chi-squared statistics with frequency tables
  with small expected cell counts.
\newblock \emph{Journal of the American Statistical Association}, 83:\penalty0
  555--560, 1988.

\bibitem[Hara and Takemura(2010)]{HT:10}
Hisayuki Hara and Akimichi Takemura.
\newblock Connecting tables with zero-one entries by a subset of a markov
  basis.
\newblock In Marlos Viana and Henry Wynn, editors, \emph{Algebraic Methods in
  Statistics and Probability II}, volume 516 of \emph{Contemporary
  Mathematics}. American Mathematical Society, 2010.

\bibitem[Hara et~al.(2012)Hara, Aoki, and Takemura]{HAT12}
Hisayuki Hara, Satoshi Aoki, and Akimichi Takemura.
\newblock Running {M}arkov chain without {M}arkov basis.
\newblock In \emph{Harmony of Gr{\"o}bner Bases and the Modern Industrial
  Society}, pages 46--62. World Scientific, 2012.

\bibitem[Hastings(1970)]{MetropolisHastingsAlgo}
W.~K. Hastings.
\newblock {M}onte {C}arlo sampling methods using {M}arkov chains and their
  applications.
\newblock \emph{Biometrika}, 57\penalty0 (1):\penalty0 97--109, 1970.
\newblock ISSN 00063444

\bibitem[Hirsch et~al.(1974)Hirsch, Devaney, and Smale]{hirsch1974differential}
M.W. Hirsch, R.L. Devaney, and S.~Smale.
\newblock \emph{Differential Equations, Dynamical Systems, and Linear Algebra}.
\newblock Pure and Applied Mathematics. Elsevier Science, 1974.
\newblock ISBN 9780080873763

\bibitem[Holzleitner et~al.(2021)Holzleitner, Gruber, Arjona-Medina,
  Brandstetter, and Hochreiter]{PPOconvergence}
Markus Holzleitner, Lukas Gruber, Jos{\'e} Arjona-Medina, Johannes
  Brandstetter, and Sepp Hochreiter.
\newblock Convergence proof for actor-critic methods applied to ppo and rudder.
\newblock In \emph{Transactions on Large-Scale Data-and Knowledge-Centered
  Systems XLVIII: Special Issue In Memory of Univ. Prof. Dr. Roland Wagner},
  pages 105--130. Springer, 2021.

\bibitem[Howard(1960)]{howard1960dynamic}
R.A. Howard.
\newblock \emph{Dynamic programming and Markov processes}.
\newblock Technology Press of Massachusetts Institute of Technology, 1960

\bibitem[Jaakkola et~al.(1994)Jaakkola, Singh, and Jordan]{RLPOMDPJaakkola}
Tommi Jaakkola, Satinder~P. Singh, and Michael~I. Jordan.
\newblock Reinforcement learning algorithm for partially observable {M}arkov
  decision problems.
\newblock In \emph{Proceedings of the 7th International Conference on Neural
  Information Processing Systems}, NIPS'94, pages 345--352, Cambridge, MA, USA,
  1994. MIT Press.

\bibitem[Ji and Jin(2016)]{JiJinAuthorsAOAS}
Pengsheng Ji and Jiashun Jin.
\newblock Coauthorship and citation networks for statisticians.
\newblock \emph{Annals of Applied Statistics}, 10\penalty0 (4):\penalty0
  1779--1812, 2016.

\bibitem[Johnson(2004)]{Johnson-BayesChiSq}
Valen~E. Johnson.
\newblock {A Bayesian $\chi^2$ test for goodness-of-fit}.
\newblock \emph{The Annals of Statistics}, 32\penalty0 (6):\penalty0 2361 --
  2384, 2004

\bibitem[Kahle et~al.(2012)Kahle, Rauh, and Sullivant]{Kahle2012POSITIVEMA}
Thomas Kahle, Johannes Rauh, and Seth Sullivant.
\newblock Positive margins and primary decomposition.
\newblock \emph{Journal of Commutative Algebra}, 6:\penalty0 173--208, 2012

\bibitem[Karwa and Petrovi\'c(2016)]{KarwaPetrovic:AOASauthors}
Vishesh Karwa and Sonja Petrovi\'c.
\newblock Coauthorship and citation networks for statisticians: Comment.
\newblock \emph{Annals of Applied Statistics}, 10\penalty0 (4):\penalty0
  1827--1834, 2016.

\bibitem[Khalil(2002)]{khalil2002nonlinear}
H.K. Khalil.
\newblock \emph{Nonlinear Systems}.
\newblock Pearson Education. Prentice Hall, 2002.
\newblock ISBN 9780130673893

\bibitem[Konda and Tsitsiklis(1999{\natexlab{a}})]{actorCriticAlgos}
Vijay Konda and John Tsitsiklis.
\newblock Actor-critic algorithms.
\newblock In S.~Solla, T.~Leen, and K.~M\"{u}ller, editors, \emph{Advances in
  Neural Information Processing Systems}, volume~12. MIT Press,
  1999{\natexlab{a}}

\bibitem[Konda and Tsitsiklis(1999{\natexlab{b}})]{Konda1999ActorCriticA}
Vijay~R. Konda and John~N. Tsitsiklis.
\newblock Actor-critic algorithms.
\newblock In \emph{NIPS}, 1999{\natexlab{b}}.

\bibitem[Lauritzen(1996)]{Lauritzen}
Steffen~L. Lauritzen.
\newblock \emph{Graphical models}.
\newblock Oxford Statistical Science Series, 1996.

\bibitem[Marbach and
  Tsitsiklis(2001)]{SimulationBasedOptimizationsMarbachTsitsiklis}
P.~Marbach and J.N. Tsitsiklis.
\newblock Simulation-based optimization of markov reward processes.
\newblock \emph{IEEE Transactions on Automatic Control}, 46\penalty0
  (2):\penalty0 191--209, 2001.
\newblock \doi{10.1109/9.905687}.

\bibitem[Metropolis and Ulam(1949)]{MonteCarloMetropolis1949}
Nicholas Metropolis and Stanislaw Ulam.
\newblock The {M}onte {C}arlo method.
\newblock \emph{Journal of the American Statistical Association}, 44\penalty0
  (247):\penalty0 335--341, 1949.

\bibitem[{Metropolis} et~al.(1953){Metropolis}, {Rosenbluth}, {Rosenbluth},
  {Teller}, and {Teller}]{MetropolisAlgorithm}
Nicholas {Metropolis}, Arianna~W. {Rosenbluth}, Marshall~N. {Rosenbluth},
  Augusta~H. {Teller}, and Edward {Teller}.
\newblock {Equation of State Calculations by Fast Computing Machines}.
\newblock \emph{Journal of Chemical Physics}, 21\penalty0 (6):\penalty0
  1087--1092, June 1953

\bibitem[Mnih et~al.(2016)Mnih, Badia, Mirza, Graves, Lillicrap, Harley,
  Silver, and Kavukcuoglu]{mnih2016asynchronous}
Volodymyr Mnih, Adria~Puigdomenech Badia, Mehdi Mirza, Alex Graves, Timothy
  Lillicrap, Tim Harley, David Silver, and Koray Kavukcuoglu.
\newblock Asynchronous methods for deep reinforcement learning.
\newblock In \emph{International conference on machine learning}, pages
  1928--1937. PMLR, 2016.

\bibitem[Petrovi\'c(2019)]{WhatIsMB}
Sonja Petrovi\'c.
\newblock What is... a {M}arkov basis?
\newblock \emph{Notices of the American Mathematical Society}, 66\penalty0
  (7):\penalty0 1088, 2019.

\bibitem[Ramaswamy and Bhatnagar(2018)]{ramaswamy2018stability}
Arunselvan Ramaswamy and Shalabh Bhatnagar.
\newblock Stability of stochastic approximations with ``controlled {M}arkov''
  noise and temporal difference learning.
\newblock \emph{IEEE Transactions on Automatic Control}, 64\penalty0
  (6):\penalty0 2614--2620, 2018.

\bibitem[Schulman et~al.(2015)Schulman, Moritz, Levine, Jordan, and
  Abbeel]{schulman2015high}
John Schulman, Philipp Moritz, Sergey Levine, Michael Jordan, and Pieter
  Abbeel.
\newblock High-dimensional continuous control using generalized advantage
  estimation.
\newblock \emph{arXiv preprint arXiv:1506.02438}, 2015.

\bibitem[Sturmfels(1996)]{St}
Bernd Sturmfels.
\newblock \emph{Gr{\"{o}}bner bases and convex polytopes}.
\newblock University Lecture Series, no. 8, American Mathematical Society,
  1996.

\bibitem[Sturmfels(2005)]{StWhatIsGB}
Bernd Sturmfels.
\newblock What is... a {G}r\"obner basis?
\newblock \emph{Notices of the American Mathematical Society}, 52\penalty0
  (10):\penalty0 1199--1200, 2005.

\bibitem[Sutton(1988)]{sutton1988learning}
Richard~S Sutton.
\newblock Learning to predict by the methods of temporal differences.
\newblock \emph{Machine learning}, 3:\penalty0 9--44, 1988.

\bibitem[Sutton and Barto(2005)]{Sutton2005ReinforcementLA}
Richard~S. Sutton and Andrew~G. Barto.
\newblock Reinforcement learning: An introduction.
\newblock \emph{IEEE Transactions on Neural Networks}, 16:\penalty0 285--286,
  2005.

\bibitem[Sutton et~al.(1999)Sutton, McAllester, Singh, and
  Mansour]{Sutton1999PolicyGM}
Richard~S. Sutton, David~A. McAllester, Satinder Singh, and Y.~Mansour.
\newblock Policy gradient methods for reinforcement learning with function
  approximation.
\newblock In \emph{NIPS}, 1999.

\bibitem[Tang et~al.(2020)Tang, Agrawal, and Faenza]{RLforIP-cut2020}
Yunhao Tang, Shipra Agrawal, and Yuri Faenza.
\newblock Reinforcement learning for integer programming: Learning to cut.
\newblock In Hal~Daum{\'e} III and Aarti Singh, editors, \emph{Proceedings of
  the 37th International Conference on Machine Learning}, volume 119 of
  \emph{Proceedings of Machine Learning Research}, pages 9367--9376. PMLR,
  13--18 Jul 2020

\bibitem[Tsitsiklis and Van~Roy(1996)]{tsitsiklis1996analysis}
John Tsitsiklis and Benjamin Van~Roy.
\newblock Analysis of temporal-diffference learning with function
  approximation.
\newblock \emph{Advances in neural information processing systems}, 9, 1996.

\bibitem[Watkins and Dayan(1992)]{WatkinsQL}
Christopher J. C.~H. Watkins and Peterl Dayan.
\newblock Q - learning.
\newblock \emph{Machine Learning}, 8\penalty0 (3):\penalty0 279---292, 1992.

\bibitem[Williams(2004{\natexlab{a}})]{Williams2004SimpleSG}
Ronald~J. Williams.
\newblock Simple statistical gradient-following algorithms for connectionist
  reinforcement learning.
\newblock \emph{Machine Learning}, 8:\penalty0 229--256, 2004{\natexlab{a}}

\bibitem[Williams(2004{\natexlab{b}})]{WilliamsREINFORCE}
Ronald~J. Williams.
\newblock Simple statistical gradient-following algorithms for connectionist
  reinforcement learning.
\newblock \emph{Machine Learning}, 8:\penalty0 229--256, 2004{\natexlab{b}}

\bibitem[Wu and De~Loera(2022)]{wu2022turning}
Yue Wu and Jes{\'u}s~A De~Loera.
\newblock Turning mathematics problems into games: Reinforcement learning and
  {G}r\"obner bases together solve integer feasibility problems.
\newblock \emph{arXiv preprint arXiv:2208.12191}, 2022.

\end{thebibliography}

\newpage



\appendix

\section{Log-linear models for categorical data}\label{appendix: log-linear models} 

Our readers should be familiar with the exponential family models from \cite{Barndorff-Nielsen}, and in particular the log-affine setup described in \cite{Lauritzen}, and may skip this section. For the general audience, we offer the following summary, extracted from the background section of \cite{MB25years}. The purpose of this section is to explain where the design matrix comes from, and how the class of log-linear models applies to any categorical data. 

Denote by $X=(X_1,\dots,X_k)$ a discrete random vector in the state space $\mathcal X=[d_1]\times\dots\times[d_k]$, where $[d]=\{1, 2, \ldots, d\}$ for any positive integer $d$. The \emph{log-affine} model is the discrete exponential family that takes the following form: 
$$f_\theta(x)= h(x) e^{\eta(\theta)^T t(x)-\psi(\theta)}.$$
In the expression above,  $t(x):\mathcal X\to\mathbb R^m$ is the sufficient statistic vector and $\eta(\theta):\Theta\to\mathbb R^m$ is the natural parameter vector. The normalizing constant, or log-partition function, is 
$\psi:\Theta\to\mathbb R$.  

One may assume that $h(x)=\bf 1$ is a constant 1 vector. This is the case in which the model is said to be \emph{log-linear}. The assumption simplifies the algebraic considerations; but more importantly, a substitution of indeterminates can easily recover the original log-affine model. 

The model representation of interest to us arises when data are arranged in a contingency table cross-classifying items according to the $k$ categories,  so that   the sufficient statistic computation  is realized as a linear operation on the table. Namely, consider $U\in \mathbb Z^{d_1\times\dots\times d_k}_{\geq 0}$, a $k$-dimensional table  
of format  $d_1\times\dots\times d_k$, whose sample space $\mathbb Z^{d_1\times\dots\times d_k}_{\geq 0}$ naturally corresponds to $\mathcal X$.  In the table $U$,  the $(i_1,\dots,i_k)$-entry $u_{i_1,\dots,i_k}$ records the number of instances of the joint event $(X_i=i_1,\dots,X_k=i_k)$ in the data.  There is a corresponding probability table  $P$, of the same format $d_1\times\dots\times d_k$, whose cells  $p_{i_1,\dots,i_k}$ are joint probabilities. 
Algebraically, since computing marginals $t(U)$ is a linear  operation on cells of the table $U$, one can also simply write $t(U)=AU$ for some integer matrix $A$ and $U$ written in vectorized form $X\in\mathbb Z^{d_1\cdots d_k}_{\geq 0}$. We use the same notation $U$ for both the matrix and table format, as the entries are identical, the vector format is just a matter of flattening the table into a vector. 
A log-linear model for $X_1,\dots,X_k$ can be equivalently expressed as 
\begin{equation}\label{log-linear model}
	P(X=x) = P(U=u) \propto \exp\{ \left<Au,\theta\right>\}, 
\end{equation}
whose sufficient statistics $t(x)$ are some set of \emph{marginals} $Au$ of the contingency table. 
Geometrically, the model is completely determined by the \emph{design matrix} $A$.

\section{Detailed comparison to related work} \label{appendix:DeLoera}

We have mentioned in the main text that  \cite{wu2022turning}  proposed casting  the general integer feasibility problem into a Markov Decision Process and applying a specific reinforcement learning algorithm to solve it.  While the problem they solve is different from the problem we solve, both rely on a particular way to sample the lattice points in a polytope. The purpose of this section is to  draw parallels and highlight differences in our work compared to  \cite{wu2022turning}, because their algorithm serves as our the starting point in terms of algorithm design. 

\cite{wu2022turning} use  two key theoretical ingredients for their game. First, using a powerful polytope universality result, they show that every integer feasibility problem is linearly equivalent to a $3$-dimensional problem where certain entries of the solution vector must be zero. 
This low-dimensional problem's feasibility and an initial solution are easily determined. Then, they propose to use RL to traverse the space of all  of its solutions  and reward the agent for discovering the states that have zeros in the entries prescribed by the linear embedding. Thus, the winning state will be a feasible solution to the original problem. 
The second  ingredient are the actions used to traverse the solution space of the $3$-dimensional problem,  known in computational algebra as a Gr\"obner basis  (see \cite{StWhatIsGB}). 
One of the main computational challenges in this setting is the fact that the low-dimensional embedding is too large to explore. Hence the application of 
RL is a logical step forward due to its ability to scale  with the problem size, as well as produce an approximate optimal answer to the problem, without searching through the entire solution space. 
In terms of RL, they use a twin delayed deep deterministic policy gradient algorithm (DDPG): 
an off-policy RL algorithm that uses two action-value networks $Q(\cdot,\cdot \ ;  \actorparam{})$ and $Q(\cdot,\cdot \ ;  \actorparam{}')$ to encode the latest optimal policy of the agent. The two $Q$ networks help with prevention of  overestimation of the Q-function during training. Furthermore, because the targets in the learning regime are Bellman updates given by the sum of the current reward and the discounted future reward, under the constantly updated policy, they are in a sense non-stationary. Hence, occasionally freezing the weights of $Q(\cdot,\cdot \ ;  \actorparam{}')$ and using it for the Bellman target computation and using $Q(\cdot,\cdot \ ;  \actorparam{})$ for storing the optimal policy helps with minimization of the loss function during training. However, such an approach requires one to keep a large enough memory buffer, as well as a burn-in sample, in order to train both networks. 
 During training, the algorithm uses samples from both the memory buffer and the burn-in sample to de-correlate the underlying dependency of the sample with respect to the current policy $\pi$. The reason is that  one desires an  i.i.d. sample during the network training stage to  obtain an unbiased estimator of the action-value function. 
They demonstrate the method on the equivalent \emph{two}-dimensional problem. 

In our algorithm, we do not use a Gr\"obner basis, but rather a lattice basis,  a subset order of magnitude smaller. In fact, this is a smallest possible Markov sub-basis, as shown in \cite[Lemma 4.2]{CDS-SISmultiwayTables2006}.  We consider the general fiber sampling problem, rather than the integer feasibility problem, and note that we can apply our algorithm to the problem they are solving, by modifying our reward function  $R_{t}$ to include a cost minimization aspect to it. 
Another key difference between our work and \cite{wu2022turning} is the RL algorithm that we implement. The off-policy approach
 will likely fail in the case when one is working with sparse fibers, where obtaining a burn-in sample using some greedy algorithm has proved a nontrivial task. 
For example,  if we naively apply a random walk algorithm to a sparse initial solution, where each move is sampled from a multinomial distribution, more often than not the random walk algorithm cannot produce a single feasible point besides the initial one (for a tiny but sparse example with low move acceptance probabilities, see \cite[Figure 1]{MB25years}). Unlike the twin delayed DDPG, we do not require a burn-in sample of observations $(s_t,a_t,r_t,s_{t+1})_i,\ i\in[N]$, as we are using an on-line training method. This approach relies on training the policy and state-value neural networks on the sample collected along the current trajectory $(s_0,a_0,r_1,s_1,a_1,r_2,\dots)$ consisting of states, actions and corresponding rewards. Furthermore, because our policy has a masking module which only uses a subset of lattice basis vectors, per a state, we are able to sample sparse fibers relatively effectively, restricting the magnitude of each move. 

\begin{figure}[H]
    \centering
    \begin{tabular}{cc}
        \subf{\includegraphics[width=70mm]{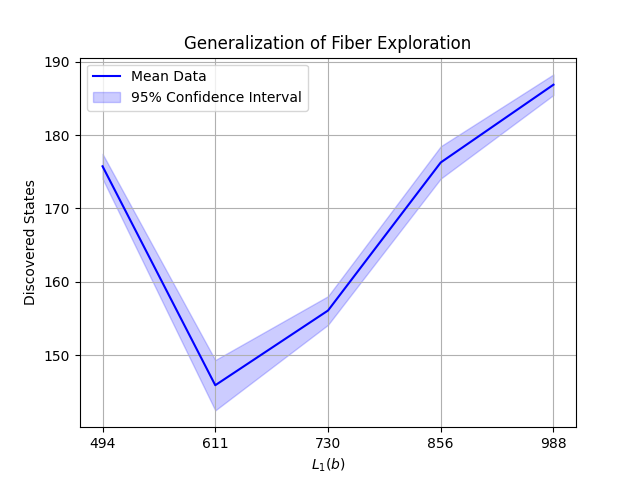}}
         {}
         \subf{\includegraphics[width=70mm]{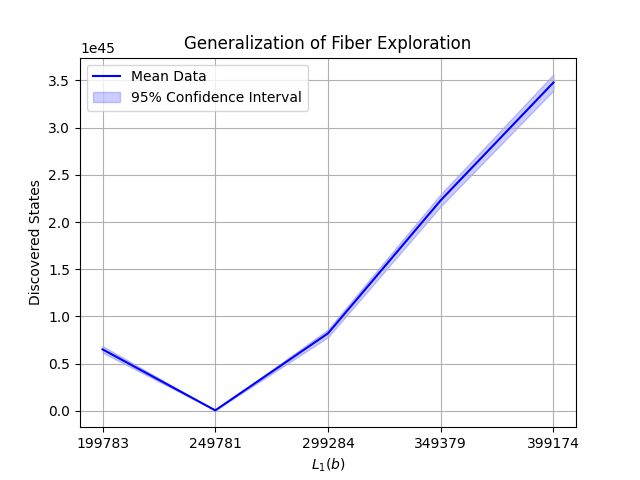}}
        {}
    \end{tabular}
    \caption{(left) We show the 95\% confidence interval of the generalization of the sampling algorithm, as we increase the margins, applied to a problem in which the initial solution is given by a graph consisted of 50 nodes. (right) We show the 95\% confidence interval of the generalization of the sampling algorithm, as we increase the margins, applied to a problem in which the initial solution is given by a graph consisted of 1000 nodes.}  
    \label{fig: generalizationSampling}
\end{figure}

Finally, we note that as in \cite[\S 6.5]{wu2022turning}, our algorithm generalizes well when perturbing the fiber by increasing the marginal vector $b$ (see Figure~\ref{fig: generalizationSampling}). Such a generalization is extremely helpful when working with limited computation resources, and it minimizes training times of the neural networks.

\section{Ergodicity of the fiber sampling MDP}\label{appendix: ergodic MDP} 

The next assumption ensures that the Markov chains obtained by following any policy $\pi$ have a stationary distributions, through which we can define loss functions in the expectation.
\begin{assumption}\label{assumption: ergodicity}
    Markov chains $\{ S_t \}$ and $\{ S_t,A_t \}$ are irreducible and aperiodic with a unique stationary probabilities $\mu(s)$ and $\eta(s,a) = \mu(s)\pi(a|s)$.
\end{assumption}
In our case the above assumption is satisfied due to the finite state space $\St$: in the problems with which we work, $P_A(b)$ is a polytope, which means it is bounded, and therefore the fiber is finite. 
Any state can be reached from any other state, in finite time. And moreover, aperiodic property is satisfied because we are able to produce a zero action $A_t = \textbf{0}$.
\cite{PPOconvergence} assume uniqueness of measure, and we follow the same setup. 

In general, the linear system $Mx=b$ may define a polyhedral system which is unbounded, in which case the state space would be infinite. Our algorithm assumes a polytope.


\section{Cumulative discounted reward}\label{appendix: cum disc reward is bdd}
 Let $G_t$ denote the  cumulative discounted reward starting at time $t$ and write
\begin{align*}
    G_t := \sum_{k=0}^\infty \gamma^k R_{t+k + 1}.
\end{align*}

Because we are sampling a finite fiber $\F{A}$, the state space is bounded, i.e. there exists $B_{\St}$ such that for all $s\in \St$, it follows that $|s|\leq B_{\St}$. Furthermore, the action space $\At$ is also bounded by design, mentioned on page \pageref{action_space_page_remark}. Therefore, we have that for all $t$, reward $R_{t+1}$ is also bounded, i.e. for all $(S_t,A_t)$ tuples, there exists a constant $B_{R}$ such that 
\begin{align*}
    |R_{t+1}|\leq B_{R}.
\end{align*}
Then, it immediately follows that 
\begin{align*}
    G_t \leq \frac{1}{1-\gamma}B_{R},\ \gamma\in (0,1)
\end{align*}
by the convergence of a geometric series. 
Let $V_{\pi}$ denote the state-value function under policy $\pi$, starting from state $S_t = s$ at time $t$, given by
\begin{align*}
    V_{\pi}(s) := \EXP_{\tau \sim \pi}\Big[ G_t \Big|\ S_t = s \Big]
\end{align*}
where $\tau$ denotes the trajectory sampled using policy $\pi$. Similarly as above, it follows that 
\begin{align*}
    V_{\pi}(s) \leq \frac{1}{1-\gamma}B_{R},\ \gamma\in (0,1).
\end{align*}
Furthermore, denote the action-value function by $Q_\pi(s,a)$ under policy $\pi$ and write
\begin{align*}
    Q_\pi(s,a):= \EXP_{\tau \sim \pi}\Big[ G_t \Big|\ S_t = s, A_t = a\Big].
\end{align*}
Again it follows that 
\begin{align*}
    Q_\pi(s,a) \leq \frac{1}{1-\gamma}B_{R},\ \gamma\in (0,1).
\end{align*}

\section{Bounded policy gradient}\label{appendix:bounded gradient}

\begin{proof}[Proof of Claim~\ref{claim: boundedGrad}]
Part (a) of the claim  follows immediately from  ergodicity, stated in Assumption \ref{assumption: ergodicity}. Let us prove part (b).  First note that for any parametarization $\actorparam{}$, the policy $\pi(a|s;\actorparam{})$ is bounded  for all $s,a$ pairs: 
\begin{align*}
    0 < \pi(a|s;\actorparam{}) \leq 1,\ ~\text{for all }\actorparam{},s,a.
\end{align*}
Furthermore, the map $\actorparam{} \mapsto \pi_{\actorparam{}}(a|s)$ is twice continuously differentiable, if we choose a smooth enough activation functions of the network $h$; for example, one can pick sigmoid, tanh, softmax, etc. Due to Assumption \ref{assumption: boundedPolicyParam}, it follows that both first and second gradients are bounded. 
\end{proof}


\section{Derivation of the actor parameter update} \label{appendix: actor parameter update}

 The objective function $L_{a}(\actorparam{})$ of the actor, under policy $ \pi_{\actorparam{}}$ parametrized by  $\actorparam{}$, 
\[
    L_{a}(\actorparam{}):= V_{\pi_{\actorparam{}}}(s_0). 
\]
This expression is exactly the state-value function starting from the initial state $S_0 = s_0$.
Next, consider the classical theorem which computes the gradient of the state-value function.

\begin{theorem}[Policy Gradient Theorem \cite{Sutton2005ReinforcementLA}] \label{thm: policyGradThm}
    Given the state-value function $V_{\pi_{\actorparam{}}}$ under parametrized policy $\pi_{\actorparam{}}$, define the gradient objective $J(\actorparam{})$ as
    \begin{align*}
        L_{a}(\actorparam{}) = V_{\pi_{\actorparam{}}}(s_0), 
    \end{align*}
    where $\mu_{\pi_{\actorparam{}}}$ is the stationary distribution of states $s$ under the Markov chain induced by policy $\pi_{\actorparam{}}$. Then, the gradient of the objective $L_{a}(\actorparam{})$ with respect to  parameter $\actorparam{}$ is given by 
    \begin{align*}
        \nabla_{\actorparam{}} L_{a}(\actorparam{}) = \nabla_{\actorparam{}} V_{\pi_{\actorparam{}}}(s_0) = \EXP_{\pi_{\actorparam{}}}\Big[ Q_{\pi_{\actorparam{}}}(S_t,A_t) \nabla_{\actorparam{}}\ln \pi_{\actorparam{}}(A_t|S_t) \Big].
    \end{align*}
\end{theorem}
\begin{proof}
Note that the following relationship holds, $Q(s,a) = R_{t+1}(s,a) + \gamma V(S_{t+1})$, because $R_{t+1}$ does not depend on $S_{t+1}$ hence it can be moved outside of the expectation. Then, given an arbitrary observed state $S_t = s$ at time $t$, we can write:  
    \begin{align*}
        &\nabla_{\actorparam{}} V_{\pi_{\actorparam{}}}(s) =\nabla_{\actorparam{}}  \EXP_{\pi_{\actorparam{}}}\Big[ G_t \Big|\ S_t = s, \pi_{\actorparam{}} \Big]\\
        &= \nabla_{\actorparam{}}  \EXP_{\pi_{\actorparam{}}}\Big[ R_{t+1}(s,A_t) + \gamma G_{t+1} \Big|\ S_t = s,\pi_{\actorparam{}} \Big] \\
        &= \nabla_{\actorparam{}}  \EXP_{\pi_{\actorparam{}}}\Big[ R_{t+1}(s,A_t) + \gamma V_{\pi_{\actorparam{}}}(S_{t+1}) \Big|\ S_t = s,\pi_{\actorparam{}} \Big] \\
        &= \nabla_{\actorparam{}} \sum_{a}\pi_{\actorparam{}}(a|s)\Big[ R_{t+1}(s,a) + \gamma V_{\pi_{\actorparam{}}}(S_{t+1})  \Big] \\
        &= \nabla_{\actorparam{}} \sum_{a}\pi_{\actorparam{}}(a|s) Q_{\pi_{\actorparam{}}}(a,s)\\
        &= \sum_{a} \Big[\nabla_{\actorparam{}} \pi_{\actorparam{}}(a|s) Q_{\pi_{\actorparam{}}}(a,s) + \pi_{\actorparam{}}(a|s)\nabla_{\actorparam{}}Q_{\pi_{\actorparam{}}}(a,s)\Big]\\
        &= \sum_{a} \Big[\nabla_{\actorparam{}} \pi_{\actorparam{}}(a|s) Q_{\pi_{\actorparam{}}}(a,s) \\
        &\hphantom{111111}+ \pi_{\actorparam{}}(a|s)\nabla_{\actorparam{}}\sum_{s',r}\Qt(s',r| s,a)(R_{t+2}+\gamma V_{\pi_{\actorparam{}}}(s'))\Big]\\
        &= \sum_{a} \Big[\nabla_{\actorparam{}} \pi_{\actorparam{}}(a|s) Q_{\pi_{\actorparam{}}}(a,s) \\
        &\hphantom{11111111111}+ \pi_{\actorparam{}}(a|s)\sum_{s'}\Qt(s'| s,a)\gamma \nabla_{\actorparam{}}V_{\pi_{\actorparam{}}}(s')\Big]\\
        &= \sum_{a} \bigg[\nabla_{\actorparam{}} \pi_{\actorparam{}}(a|s) Q_{\pi_{\actorparam{}}}(a,s) + \pi_{\actorparam{}}(a|s)\sum_{s'}\Qt(s'| s,a)\\\
        &\hphantom{11111111111}\cdot\gamma \sum_{a'}\Big[\nabla_{\actorparam{}} \pi_{\actorparam{}}(a'|s') Q_{\pi_{\actorparam{}}}(a',s') \\
        &\hphantom{11111111111}+ \pi_{\actorparam{}}(a'|s')\sum_{s''}\Qt(s''| s',a')\gamma \nabla_{\actorparam{}}V_{\pi_{\actorparam{}}}(s'')\Big]\bigg]\\
        &= \sum_{x\in \St} \sum_{k=0}^\infty\gamma^k Pr(s\rightarrow x, k, \pi_{\actorparam{}})\sum_{a}\nabla_{\actorparam{}}\pi_{\actorparam{}}(a|x)Q_{\pi_{\actorparam{}}}(a,x).
    \end{align*}
Here $ Pr(s\rightarrow x, k, \pi_{\actorparam{}})$ is the probability of being in state $x$ after $k$ steps under policy $\pi_{\actorparam{}}$ and starting from state $s$. Note that the discounting factors $\gamma, \gamma^2,...$ have been absorbed into the probability. It follows then that 
\begin{align*}
    &\nabla_{\actorparam{}}  
    L_{a}(\actorparam{}) =\nabla_{\actorparam{}} V_{\pi_{\actorparam{}}}(s_0)\\
    &= \sum_{s} \sum_{k=0}^\infty Pr(s_0\rightarrow s, k, \pi_{\actorparam{}})\sum_{a}\nabla_{\actorparam{}}\pi_{\actorparam{}}(a|s )Q_{\pi_{\actorparam{}}}(a,s)\\
    &= \sum_{s}\eta(s)\sum_{a}\nabla_{\actorparam{}}\pi_{\actorparam{}}(a|s )Q_{\pi_{\actorparam{}}}(a,s)\\
    &= \sum_{s'}\eta(s')\sum_{s}\frac{\eta(s)}{ \sum_{s'}\eta(s')}\sum_{a}\nabla_{\actorparam{}}\pi_{\actorparam{}}(a|s )Q_{\pi_{\actorparam{}}}(a,s)\\
    &= \sum_{s'}\eta(s')\sum_{s}\mu_{\pi_{\actorparam{}}}(s)\sum_{a}\nabla_{\actorparam{}}\pi_{\actorparam{}}(a|s )Q_{\pi_{\actorparam{}}}(a,s)\\
    &\propto \sum_{s}\mu_{\pi_{\actorparam{}}}(s)\sum_{a}\nabla_{\actorparam{}}\pi_{\actorparam{}}(a|s )Q_{\pi_{\actorparam{}}}(a,s),
\end{align*}
where $\mu_{\pi_{\actorparam{}}}(s)$ is the on-policy, stationary state distribution under policy $\pi_{\actorparam{}}$. In the case of a continuing task, the proportionality constant is $1$, therefore 
\begin{align*}
    \nabla_{\actorparam{}} L_{a}(\actorparam{}) 
    &= \sum_{s}\mu_{\pi_{\actorparam{}}}(s)\sum_{a}\nabla_{\actorparam{}}\pi_{\actorparam{}}(a|s )Q_{\pi_{\actorparam{}}}(s,a).
\end{align*}
We can further manipulate the expression above to get a gradient computation found in the REINFORCE algorithm \cite{WilliamsREINFORCE}:
\begin{align*}
    \nabla_{\actorparam{}} L_{a}(\actorparam{}) &= \sum_{s}\mu_{\pi_{\actorparam{}}}(s)\sum_{a}\nabla_{\actorparam{}}\pi_{\actorparam{}}(a|s )Q_{\pi_{\actorparam{}}}(s,a)\\
    &=  \EXP_{\pi_{\actorparam{}}}\Big[\sum_{a}\nabla_{\actorparam{}}\pi_{\actorparam{}}(a|S_t )Q_{\pi_{\actorparam{}}}(S_t,a)\Big]\\
    &= \EXP_{\pi_{\actorparam{}}}\Big[\sum_{a}\pi_{\actorparam{}}(a|S_t )Q_{\pi_{\actorparam{}}}(S_t,a)\frac{\nabla_{\actorparam{}}\pi_{\actorparam{}}(a|S_t)}{\pi_{\actorparam{}}(a|S_t)}\Big]\\
    &= \EXP_{\pi_{\actorparam{}}}\Big[Q_{\pi_{\actorparam{}}}(S_t,A_t)\frac{\nabla_{\actorparam{}}\pi_{\actorparam{}}(A_t|S_t)}{\pi_{\actorparam{}}(A_t|S_t)}\Big]\\
    &= \EXP_{\pi_{\actorparam{}}}\Big[Q_{\pi_{\actorparam{}}}(S_t,A_t)\nabla_{\actorparam{}}\ln\pi_{\actorparam{}}(A_t|S_t)\Big].
\end{align*}
\end{proof}

We can modify the gradient computation above in a way that the value of the gradient remains the same, while the variance of an estimator of this gradient is reduced. We accomplish this by introducing a baseline function, denoted $b(\cdot)$. The following Lemma \ref{lemma: baseline} shows that this modification indeed does not change the policy gradient.
\begin{lemma}\label{lemma: baseline}
    Given the baseline function $b(S_t)$, we can write the gradient of the state-value function in the form of 
    \begin{align*}
        \nabla_{\actorparam{}}L_{a}(\actorparam{}) &= \EXP_{\pi_{\actorparam{}}}\Big[\big(Q_{\pi_{\actorparam{}}}(S_t,A_t)-b(S_t)\big)\nabla_{\actorparam{}}\ln\pi_{\actorparam{}}(A_t|S_t)\Big].
    \end{align*}
\end{lemma}
\begin{proof}
    We want to show that 
    \begin{align*}
        \EXP_{\pi_{\actorparam{}}}\Big[b(S_t)\nabla_{\actorparam{}}\ln\pi_{\actorparam{}}(A_t|S_t)\Big] = 0, 
    \end{align*}
    from which the result will follow. We note that one can rewrite the expectation as:  
    \begin{align*}
         &\EXP_{\pi_{\actorparam{}}}\Big[b(S_t)\nabla_{\actorparam{}}\ln\pi_{\actorparam{}}(A_t|S_t)\Big] \\ &\hphantom{1111111111111111}= \sum_{S_t=s}\mu_{\pi_{\actorparam{}}}(s)\sum_{A_t=a}\nabla_{\actorparam{}}\pi_{\actorparam{}}(a|s )b(s) \\
         &\hphantom{1111111111111111}= \sum_{S_t=s}\mu_{\pi_{\actorparam{}}}(s)b(s)\nabla_{\actorparam{}}\sum_{A_t=a}\pi_{\actorparam{}}(a|s ) \\
         &\hphantom{1111111111111111}=\sum_{S_t=s}\mu_{\pi_{\actorparam{}}}(s)b(s)\nabla_{\actorparam{}}1 = 0.
    \end{align*}
    Thus, it follows that 
    \begin{align*}
        \nabla_{\actorparam{}} L_{a}(\actorparam{}) &= \EXP_{\pi_{\actorparam{}}}\Big[\big(Q_{\pi_{\actorparam{}}}(S_t,A_t)-b(S_t)\big)\nabla_{\actorparam{}}\ln\pi_{\actorparam{}}(A_t|S_t)\Big] \\
        &=\nabla_{\actorparam{}} L_{a}(\actorparam{}) - 0 = \nabla_{\actorparam{}} L_{a}(\actorparam{}).
    \end{align*}
\end{proof}
Therefore, as long as the baseline function does not depend on the action $A_t = a$, we can absorb it inside of the gradient computation. For example, a popular choice is $b(S_t) = V_{\pi_{\actorparam{}}}(S_t)$. Then, we get 
\begin{align*}
    A_{\actorparam{}}(S_t,A_t) = Q_{\pi_{\actorparam{}}}(S_t,A_t) - V_{\pi_{\actorparam{}}}(S_t),
\end{align*}
which is called the advantage function \cite{baird1993advantage}.
 Then the gradient is written as 
\begin{align}\label{eqn: TDgradient}
    \nabla_{\actorparam{}} L_{a}(\actorparam{}) = \EXP_{\pi_{\actorparam{}}}\Big[A_{\actorparam{}}(S_t,A_t)\nabla_{\actorparam{}}\ln\pi_{\actorparam{}}(A_t|S_t)\Big].
\end{align}
Notice that we can approximate the gradient in \eqref{eqn: TDgradient} by defining an estimator 
\begin{align*}
    \widehat{\nabla_{\actorparam{}} L_{a}(\actorparam{})} := \frac{1}{K}\sum_{l=0}^{K-1}\delta_{\actorparam{}}(l)\nabla_{\actorparam{}}\ln\pi_{\actorparam{}}(A_l|S_l), 
\end{align*}
where
\begin{align*}
    \delta_{\actorparam{}}(l) := R_{l+1}(S_l,A_l) + \gamma V_{\pi_{\actorparam{}}}(S_l+A_l) - V_{\pi_{\actorparam{}}}(S_l)
\end{align*}
is called a temporal difference (TD) \cite{sutton1988learning}. Note that $S_{l+1} = S_l + A_l$ due to the environment dynamics defined in \ref{defn:MDP}. 
We can go a step further and generalize the estimate above.
\begin{definition}[Generalized Advantage Estimate (GAE) \cite{schulman2015high}]\label{def: GAE}
    Define the advantage estimate in the following way
    \begin{align*}
        \hat{A}^{(1)}(t) &= R_{t+1} + \gamma V(S_{t+1}) - V(S_t) = \delta(t)\\
        \hat{A}^{(2)}(t) &= R_{t+1} + \gamma R_{t+2} + \gamma^2 V(S_{t+2}) - V(S_t)\\
        &= \delta(t) + \gamma(\delta(t+1))\\
        &\vdots\\
        \hat{A}^{(k)}(t) &= \sum_{l=0}^k \gamma^{l}\delta(t+l)\\
        &\vdots\\
        \hat{A}^{(\infty)}(t) &= \sum_{l=0}^\infty \gamma^{l}\delta(t+l) = - V(S_t) + \sum_{l=0}^\infty\gamma^{l}R_{t+1+l}.
    \end{align*}
    Then, for $\lambda\in [0,1]$ and policy $\pi_{\actorparam{}}$, define the Generalized Advantage Estimate as
    \begin{align*}
        \widehat{A_{GAE}}(t) = (1-\lambda)\sum_{k=0}^\infty \lambda^{k}\hat{A}^{(k)}(t) = \sum_{l=0}^\infty (\gamma \lambda)^{l}\delta(t+l). 
    \end{align*}
\end{definition}
Note that the GAE is a biased estimator of the advantage function $A_{\actorparam{}}(S_t,A_t) = Q_{\pi_{\actorparam{}}}(S_t,A_t) - V_{\pi_{\actorparam{}}}(S_t)$, in which the bias comes from the discount factor $\lambda$. It decomposes into an estimator of $Q_{\pi_{\actorparam{}}}(S_t,A_t)$ and the baseline function $b(S_t)$. We thus  define a generalized advantage estimator policy gradient as: 
\begin{align*}
    \nabla_{\actorparam{}} L_{GAE}(\actorparam{}) := \EXP_{\pi_{\actorparam{}}}\Big[\widehat{A_{GAE}}(t)\nabla_{\actorparam{}}\ln\pi_{\actorparam{}}(A_t|S_t)\Big].
\end{align*}
Furthermore, note that because the TDs are bounded, due to the state-value function and instantaneous rewards being bounded, the $\widehat{A_{GAE}}$ is bounded as well for all $t$. From the previous subsection we know that $\nabla_{\actorparam{}}\ln\pi_{\actorparam{}}(A_t|S_t)$ is bounded as well and thus, by the Fubini-Tonelli theorem, it follows that 
\begin{align*}
    \nabla_{\actorparam{}} L_{GAE}(\actorparam{}) &= \EXP_{\pi_{\actorparam{}}}\Big[\widehat{A_{GAE}}(t)\nabla_{\actorparam{}}\ln\pi_{\actorparam{}}(A_t|S_t)\Big]\\
    &= \sum_{l=0}^\infty (\gamma \lambda)^l \EXP_{\pi_{\actorparam{}}}\Big[\delta(t+l)\nabla_{\actorparam{}}\ln\pi_{\actorparam{}}(A_t|S_t)\Big].
\end{align*}
Then we have that 
\begin{align*}
    \nabla_{\actorparam{}} L_{a}(\actorparam{}) &=  \EXP_{\pi_{\actorparam{}}}\Big[A_{\actorparam{}}(S_t,A_t)\nabla_{\actorparam{}}\ln\pi_{\actorparam{}}(A_t|S_t)\Big]\\
    &\approx 
 \nabla_{\actorparam{}} L_{GAE}(\actorparam{})
\end{align*}
where equality holds when $\lambda = 1$.  Note that the TD terms $\delta(k+l)$ are exponentially discounted and thus terms further in the future contribute significantly less to the value of the infinite sum. Hence, we can truncate the sum up to some step $K-1$ such that
\begin{align}\label{eqn: approxGrad}
     \actorgrad(\actorparam{}) &:=  \sum_{l=0}^{K-1}(\gamma \lambda)^l \EXP_{\pi_{\actorparam{}}}\Big[\delta(t+l)\nabla_{\actorparam{}}\ln\pi_{\actorparam{}}(A_t|S_t)\Big]\\
     &= \sum_{l=0}^{K-1}(\gamma \lambda)^l \Big(\EXP_{\pi_{\actorparam{}}}\Big[R_{t+1}(S_t,A_t)\nabla_{\actorparam{}}\ln\pi_{\actorparam{}}(A_t|S_t)\Big]\nonumber\\
     &+ \EXP_{\pi_{\actorparam{}}}\Big[\gamma V_{\pi_{\actorparam{}}}(S_{t+1})\nabla_{\actorparam{}}\ln\pi_{\actorparam{}}(A_t|S_t)\Big]\nonumber \\
     &-  \EXP_{\pi_{\actorparam{}}}\Big[V_{\pi_{\actorparam{}}}(S_{t})\nabla_{\actorparam{}}\ln\pi_{\actorparam{}}(A_t|S_t)\Big] \Big)\nonumber.
\end{align}
All the above computations have been derived under the assumption that we have an exact expression of the state-value function $V_{\pi_{\actorparam{}}}$. However, in practice, this is usually not the case and we can only work with an approximation of such function under some parametarization $\criticparam{}$. In what follows, we show that such parametarization does not pose any issue in the gradient estimate. We follow the work of \cite{actorCriticAlgos} and \cite{tsitsiklis1996analysis} and define, for any $\actorparam{}$, two inner products $\langle \cdot, \cdot \rangle_{\theta}^1$ and $\langle \cdot, \cdot \rangle_{\theta}^2$ of two pairs of real-valued functions $F_1^1,F_2^1$, on $\mathcal{S}\times \mathcal{A}$, and $F_1^2,F_2^2$ on $\mathcal{S}$ viewed as vectors in $\mathbb{Z}^{|\mathcal{S}||\mathcal{A}|}$ and $\mathbb{Z}^{|\mathcal{S}|}$, by 
\begin{align*}
    &\langle F_1^1, F_2^1 \rangle_{\theta}^1 := \sum_{s,a}\mu_{\theta{}}(s)\pi_{\theta}(a|s) F_1^1(s,a)F_2^1(s,a)\\
    &\langle F_1^2, F_2^2 \rangle_{\theta}^2 := \sum_{s}\mu_{\theta{}}(s)F_1^2(s)F_2^2(s).
\end{align*}
 If we only consider partial derivative w.r.t parameter  components $\actorparam{}^i$ we can write equation \eqref{eqn: approxGrad} as
\begin{align*}
    \frac{\partial}{\partial \actorparam{}^i} \actorgrad(\actorparam{}) &= \sum_{l=0}^{K-1}(\gamma \lambda)^l \Big(\innerProd{R_{t+1}(S_t,A_t)}{ \psi_{\actorparam{}}^{1,i}}{\actorparam{}}^1\\
     &+ \innerProd{V_{\pi_{\actorparam{}}}(S_{t+1})}{ \psi_{\actorparam{}}^{2,i}}{\actorparam{}}^2 - \innerProd{V_{\pi_{\actorparam{}}}(S_{t})}{ \psi_{\actorparam{}}^{2,i}}{\actorparam{}}^2 \Big)\\
     &\ ~i = 1,\dots, n, 
\end{align*}
where $\psi_{\actorparam{}}^{1,i}$, $\psi_{\actorparam{}}^{2,i}$ denote the $i-th$ vector components of 
\begin{align*}
    &\psi_{\actorparam{}}^1(S_t,A_t) := \nabla_{\actorparam{}}\ln \pi_{\actorparam{}}(A_t|S_t)\\
    &\psi_{\actorparam{}}^2(S_t) := \sum_{a}\pi_{\actorparam{}}(a|S_t)\nabla_{\actorparam{}}\ln \pi_{\actorparam{}}(a|S_t).
\end{align*} 
To clarify the last sentence, consider the objects $\psi_{\actorparam{}}(S_t,A_t)^1$ and $\psi_{\actorparam{}}(S_t)^2$, which are already vectors of a fixed state-action $(S_t,A_t)$ and state $S_t$ respectively, due to the gradient term inside it. However, in the view of the inner products $\innerProd{\cdot}{\cdot}{\actorparam{}}^j$, $j=1,2$, each component is a different state-action $(s,a)$ and state $s$ and not the partial derivatives of the gradient itself. Thus, we are looking at a vector with components being vectors as well. The outer vector is comprised of state-action components $S_t=s,A_t=a$ and state components $S_t=s$ and the inner vectors are comprised of the gradients with respect to $\actorparam{}$. Hence, if we only consider the component of the gradient indexed by the partial derivative $ \frac{\partial}{\partial \actorparam{}^i}$, we get vectors $\psi_{\actorparam{}}^{1,i}$ in $\mathbb{Z}^{|\St||\At|}$ and $\psi_{\actorparam{}}^{2,i}$ in $\mathbb{Z}^{|\St|}$.

Furthermore, define projections $\Pi^1_{\actorparam{}}: \mathbb{Z}^{|\St||\At|} \mapsto \Psi_{\actorparam{}}^1$, $\Pi^2_{\actorparam{}}: \mathbb{Z}^{|\St|} \mapsto \Psi_{\actorparam{}}^2$ as $\Pi^1_{\actorparam{}} F := \arg\min_{\hat{F}\in \Psi_{\actorparam{}}^1}\norm{F - \hat{F}}{\actorparam{}}^1$ and $\Pi^2_{\actorparam{}} F := \arg\min_{\hat{F}\in \Psi_{\actorparam{}}^2}\norm{F - \hat{F}}{\actorparam{}}^2$, respectively. 
Here, $\norm{\cdot}{\actorparam{}}^1$ and $\norm{\cdot}{\actorparam{}}^2$ are the norms on $\mathbb{Z}^{|\St||\At|}$,$\mathbb{Z}^{|\St|}$ induced by inner products $\innerProd{\cdot}{\cdot}{\actorparam{}}^1$ and $\innerProd{\cdot}{\cdot}{\actorparam{}}^2$. By definition of a projection, it follows that the error of the projection is orthogonal to the projection space. Hence, in both cases, $\innerProd{\Pi^1_{\actorparam{}}R_{t+1}(S_t,A_t) - R_{t+1}(S_t,A_t) }{\psi_{\actorparam{}}^{1,i}}{\actorparam{}}^1 = 0$ and   $\innerProd{\Pi^2_{\actorparam{}}V_{\pi_{\actorparam{}}} - V_{\pi_{\actorparam{}}} }{\psi_{\actorparam{}}^{2,i}}{\actorparam{}}^2 = 0$ imply that $\innerProd{\Pi^1_{\actorparam{}}R_{t+1}(S_t,A_t)}{\psi_{\actorparam{}}^{1,i}}{\actorparam{}}^1 = \innerProd{R_{t+1}(S_t,A_t)}{\psi_{\actorparam{}}^{1,i}}{\actorparam{}}^1$ and $\innerProd{\Pi^2_{\actorparam{}}V_{\pi_{\actorparam{}}}}{\psi_{\actorparam{}}^{2,i}}{\actorparam{}}^2 = \innerProd{V_{\pi_{\actorparam{}}}}{\psi_{\actorparam{}}^{2,i}}{\actorparam{}}^2$
which in turn shows that it suffices to learn the projection of $V_{\pi_{\actorparam{}}}$ onto $\Psi_{\actorparam{}}$ in order to compute the approximation of the gradient $\actorgrad(\actorparam{})$.
So, let $V_{\criticparam{}}$ be a projection of the state-value function onto the subspace spanned by $\Psi_{\actorparam{}}$, and let 
\begin{align*}
    \delta_{\criticparam{}}(t) := R_{t+1} + \gamma V_{\criticparam{}}(t+1) - V_{\criticparam{}}(t).
\end{align*}
Then we write the gradient approximation of the exact gradient as 
\begin{align}
     \actorgrad(\actorparam{},\criticparam{}) := \sum_{l=0}^{K-1}(\gamma \lambda)^l \EXP_{\pi_{\actorparam{}}}\Big[\delta_{\criticparam{}}(t+l)\nabla_{\actorparam{}}\ln\pi_{\actorparam{}}(A_t|S_t)\Big].
\end{align}
Furthermore, because we are collecting data in a on-line fashion, we are estimating the gradient along a truncated trajectory $\tau_{t:t+K}$. Hence, the gradient estimator can be computed as 
\begin{align}
    \actorgradestim(\actorparam{},\criticparam{}) := \frac{1}{K}\sum_{k=0}^{K-1}\nabla_{\actorparam{}}\ln\pi_{\actorparam{}}(A_k|S_k)\sum_{l=k}^{K-1} (\gamma \lambda)^{l-k}\delta_{\criticparam{}}(t+l) .
\end{align}

These detailed considerations allow us to define the actor parameter update as in Definition~\ref{def: actorUpdate}.


\section{Feature vector bounds}\label{appendix: feature vector}
In Section~\ref{section: critic}, we noted that, not having access to the analytical form of the state-value function $V$, we parametrize it:  
\[
    V_{\criticparam{}}(S_t) := \phi_{\actorparam{}}(S_t)^T\criticparam{}\approx V_{\actorparam{}}.
\]
This parametrization can be written in matrix form as follows: 
\begin{align*}
    V_{\criticparam{}}(S_t) &= 
    \begin{bmatrix}
    &\phi_{\actorparam{}}(S_1)\\
    &\vdots\\
    &\phi_{\actorparam{}}(S_n)
    \end{bmatrix}\criticparam{}\\
    &= 
    \begin{bmatrix}
    &\phi_{1,\actorparam{}}(S_1) & \dots & \phi_{n,\actorparam{}}(S_1)\\
    &\vdots\\
    &\phi_{1,\actorparam{}}(S_n)& \dots & \phi_{n,\actorparam{}}(S_n)
    \end{bmatrix}\criticparam{} = \Phi \criticparam{}.
\end{align*}
\begin{claim}\label{claim: featureVector}
Adopting the notation above, the following conditions are satisfied: 

  (a) For every $s\in \St$, the mapping $\actorparam{} \mapsto \phi_{\actorparam{}}$ is continuous, bounded, and differentiable.

  (b) The span of vectors $\phi_{\actorparam{}}^j$, $j=1,\dots, m$ in $\mathbb{Z}^{|\St|}$, denoted by $\Phi_{\actorparam{}}$ contains $\Psi_{\actorparam{}}$.
\end{claim}
\begin{proof}[Proof of Claim~\ref{claim: featureVector}]
Part $(a)$ of the claim follows by assumption \ref{assumption: boundedPolicyParam} and Claim \ref{claim: boundedGrad}.

 Part $(b)$ is a technicality needed to ensure that the neural network is well posed. Essentially, it says the following. Given the feature vector space $\Phi_{\actorparam{}}$, the policy network maps the vector space to another vector space defined implicitly by $\pi_{\actorparam{}}(a|\phi_{\actorparam{}}(S_t))$. Taking the logarithm and then back-propagating with a gradient $\nabla_{\actorparam{}}$, we again end up in the feature space $\Phi_{\actorparam{}}$. Thus the assumption ensures that taking the gradient of the network, we will end back in the feature space. Such an assumption is naturally satisfied with a proper choice of activation functions. It also ensures that the projector $\Pi_{\actorparam{}}$ does not become ill-conditioned. 
\end{proof} 


\section{Critic minimization objective and parameter update} \label{appendix: critic}

As mentioned on page~\pageref{section: critic}, one of the most basic but well established minimization objectives, the $n$-step bootstrap, minimizes the difference between collected rewards along a truncated trajectory $\tau_{t:t+K}$ and the parametrized state-value function $V_{\criticparam{}}$. 

In mathematical detail, consider the  difference between collected rewards and state-value function: 
\begin{align}
   &L_{\criticgrad}(\actorparam{},\criticparam{}) =\nonumber \\
    &\min_{\criticparam{}} \frac{1}{2}\EXP_{\pi_{\actorparam{}}}\Big[\sum_{k=0}^{K-1}\Big( - V_{\criticparam{}}(S_{t+k})\nonumber\\
    &+ \big(\sum_{j=k}^{K-1} \gamma^{j-k} R_{t+1+j}(S_{t+j},A_{t+j})  +\gamma^{K-k} V_{\criticparam{}}(S_{t+K})\big) 
 \Big)^2\Big].
\end{align}


Now, consider the definition of critic parameter update in~\ref{def: criticUpdate}. 
Assume that the system is in time $t+K-1$ and take the gradient of  minimization objective~\eqref{eqn: critic minimization objective} with respect to the parameter $\criticparam{t+K-1}$: 
\begin{align}\label{eqn: criticGradEstim}
    &\criticgrad(\actorparam{t+K-1},\criticparam{t+K-1}) :=\nonumber \\
    &\EXP_{\pi_{\actorparam{}}}\Big[\sum_{k=0}^{K-1} \Big(V_{\criticparam{t+K-1}}(S_{t+k})\nonumber\\
    &- \big(\sum_{j=k}^{K-1} \gamma^{j-k} R_{t+1+j} + \gamma^{K-k}V_{\criticparam{t+K-1}}(S_{t+K})\big)\Big)\nonumber\\
    &\cdot\Big(\phi_{\actorparam{t+K-1}}(S_{t+k}) - \gamma^{K-k}\phi_{\actorparam{t+K-1}}(S_{t+K})\Big)\Big]
\end{align}
which in turn gives us the critic updated at time $t+K-1$:
\begin{align}
    &\criticparam{t+K} = \criticparam{t+K-1}+ \beta(t+K-1) \nonumber \\
    &\cdot \sum_{k=0}^{K-1}\Big( V_{\criticparam{t+K-1}}(S_{t+k})- \big(\sum_{j=k}^{K-1} \gamma^{j-k} R_{t+1+j}(S_{t+j},A_{t+j})  \nonumber\\
    &+\gamma^{K-k} V_{\criticparam{t-K+1}}(S_{t+K})\big)\Big)\nonumber\\
    &\cdot\Big(\phi_{\actorparam{t+K-1}}(S_{t+k}) - \gamma^{K-k}\phi_{\actorparam{t+K-1}}(S_{t+K}) \Big)\label{eqn: a2cScheme2},\\
    & \nabla_{\criticparam{t+K-1}}V_{\criticparam{t+K-1}}(S_{t+l}) = \phi_{\actorparam{t+K-1}}(S_{t+l})\nonumber.
\end{align}    



\section{Convergence: proof of Theorem~\ref{thm: main}}\label{appendix: convergence_proof} 
We aim to setup our algorithm in the framework presented by \cite{BORKAR1997291} and apply their convergence theorem. 
\label{first page of proof} 
In the most general case, we are presented with the following stochastic iterative scheme 
\begin{align}
    x_{n+1} &= x_{n} + a(n)[h(x_n,y_n) + M_{n+1}^{(1)}]\label{eqn: originalscheme1} \\
    y_{n+1} &= y_{n} + b(n)[g(x_n,y_n) + M_{n+1}^{(2)}]\label{eqn: originalscheme2}
\end{align}
where $a(n)$ and $b(n)$ are step-sizes for the iterates, $h$ and $g$ are Lipschitz continues functions of the parameters at hand, and $M_{n+1}^{(i)}$ is a martingale difference with bounded second moments for $i=1,2$. The difficulty of showing convergence of such a scheme is that both equations above are coupled with respect to to the parameters they are updating. One potential approach would be to fix one parameter and converge the second, and then converge the first parameter as well. Although somewhat viable, the approximate convergence times for such approach would be much larger than what the o.d.e. method proposes. Namely, in this method the step-sizes $a(n)$ and $b(n)$ are chosen in such a way that one parameter is updated with a much slower rate than the other. In that case, the slower iterate (parameter) becomes quasi-stationary from the perspective of the fast iterate. By picking such viable step-sizes, one can show that the stochastic iterative scheme given by \eqref{eqn: originalscheme1} and \eqref{eqn: originalscheme2} converges to a local optimum $(x^*,y^*)$. The idea behind introducing ordinary differential equations in the analysis of the above scheme is that one can show that both iterates in \eqref{eqn: originalscheme1} and \eqref{eqn: originalscheme2} asymptotically follow the trajectories of certain o.d.e.'s. If these o.d.e.'s have an attractor (an equilibrium point), then the scheme converges to the same point.

The proof heavily leverages the setup presented above as well as the derivations in \cite{actorCriticAlgos}. Before presenting the proof, we need to re-cast our algorithm in the form of Equtions~\eqref{eqn: originalscheme1} and~\eqref{eqn: originalscheme2} and check whether it satisfies the conditions presented above. 

To this end, let $\actorparam{t}$ and $\criticparam{t}$ denote the actor and critic parameters when the system is in time $t$, respectively. 
Our goal is to show that the parameters $(\actorparam{t},\criticparam{t}) \longrightarrow (\actorparam{}^*,\criticparam{}^*)$ converge to a global optimum $(\actorparam{}^*,\criticparam{}^*)$. We do this by showing that iterates in Equations~\eqref{eqn: a2cScheme1} and~\eqref{eqn: a2cScheme2} satisfy conditions \ref{condition: C1}, \ref{condition: C2}, \ref{condition: C3}, \ref{condition: C4}, \ref{condition: C5}, \ref{condition: C6}, and then we apply Lemma \ref{lemma: convergence2scale} and Theorem \ref{thm: convergence2scale}. 

We first put Equations \eqref{eqn: a2cScheme1} and \eqref{eqn: a2cScheme2} in the correct form. Let 
\begin{align}
    M_{t+1}^{(c)} = \criticgradestim(\actorparam{t},\criticparam{t}) - \criticgrad(\actorparam{t},\criticparam{t}),\\
     M_{t+1}^{(a)} = \actorgradestim(\actorparam{t},\criticparam{t}) - \actorgrad(\actorparam{t},\criticparam{t}). 
\end{align}
Then the iterates above assume the following form
\begin{align}
    \criticparam{t+1} &= \criticparam{t} + \beta(t)[\criticgrad(\actorparam{t},\criticparam{t}) + M_{t+1}^{(c)}]\label{eqn: moddedA2CScheme1} \\
    \actorparam{t+1} &= \actorparam{t} + \alpha(t)[\actorgrad(\actorparam{t},\criticparam{t}) + M_{t+1}^{(a)}]\label{eqn: moddedA2CScheme2}
\end{align}
where equations \eqref{eqn: moddedA2CScheme1},\eqref{eqn: moddedA2CScheme2} are now in the form of \eqref{eqn: originalscheme1} and \eqref{eqn: originalscheme2},  respectively. 
 In order for the Theorem 1 in \cite{BORKAR1997291} (also found in \cite{borkar2009stochastic}) to be satisfied, we need to show that our algorithm satisfies the following conditions. 

\begin{condition}\label{condition: C1}
    Functions $h:\mathbb{R}^{d+k} \longrightarrow \mathbb{R}^d$ and $g:\mathbb{R}^{d+k} \longrightarrow \mathbb{R}^k$ are continues and Lipschitz with respect to both input parameters.
\end{condition}

\begin{condition}\label{condition: C2}
    Objects $M_{n+1}^{(i)}$, $i=1,2$ are martingale differences with bounded second moments, i.e.
    \begin{align*}
        \EXP[M_{n+1}^{(i)}|\filtration{n}] = 0,\ ~i=1,2
    \end{align*}
    where $\filtration{n} = \sigma(x_m,y_m,M_{m}^{(1)},M_{m}^{(2)}, m\leq n)$, $\filtration{0}\subset\dots \subset\filtration{n}\subset \filtration{n+1}\subset \dots \subset \filtration{}$ is the filtration generated by information at subsequent times, consisted of prameters $x_n,y_n$ and martingale differences $M_{m}^{(i)},\ ~i=1,2$. Moreover their second moments are bounded, i.e.
    \begin{align*}
        \EXP[\norm{M_{n+1}^{(i)}}{}^2\ |\filtration{n}] \leq K(1+\norm{x_n}{}^2 + \norm{y_n}{}^2),\ ~i=1,2
    \end{align*}
    for a suitable constant $K$.
\end{condition}

\begin{condition}\label{condition: C3}
    Step-sizes $\{a(n)\}$ and $\{b(n)\}$ are positive scalars satisfying Robbins–Monro criteria
    \begin{align*}
        \sum_{n}a(n) = \sum_{n}b(n) = \infty,\ ~\sum_{n}(a(n)^2&+b(n)^2) < \infty,\\
        &~\frac{b(n)}{a(n)} \longrightarrow 0
    \end{align*}
\end{condition}
In the  above, $\frac{b(n)}{a(n)} \longrightarrow 0$ implies that the sequence $b(n) \longrightarrow 0$ converges to 0 faster compared to the sequence $\{a(n)\}$ and so the iterate in Equation~\eqref{eqn: originalscheme2} operates on a slower timescale compared to the iterate in Equation~\eqref{eqn: originalscheme1}. Next, consider an o.d.e. defined as 
\begin{align}\label{eqn: ode1}
    \dot{x}(t) = h(x(t),y)
\end{align}
where $y$ is being held fixed and treated as a constant parameter. 

\begin{condition}\label{condition: C4}
    The o.d.e. in Equation~\eqref{eqn: ode1} has a globally asymptotically stable equilibrium $\lambda(y)$, where $\lambda: \mathbb{R}^k \longrightarrow \mathbb{R}^d$ is a Lipschitz map.
\end{condition}
Given a small $\epsilon$, this condition tells us that the solution $x(t)$ tracks $\lambda(y(t))$ for $t>0$. This gives rise to the following o.d.e.: 
\begin{align}\label{eqn: ode2}
    \dot{y}(t) = h(\lambda(y(t)),y(t)), 
\end{align}
where now the equation solely depends on trajectory $y(t)$. Finally, equation \eqref{eqn: ode2} yields another condition: 
\begin{condition}\label{condition: C5}
    The o.d.e. \eqref{eqn: ode2} has a globally asymptotically stable equilibrium $y^*$.
\end{condition}
Given the last two conditions, we expect the trajectory $(x(t),y(t))$ to approximately converge to the point $(\lambda(y^*), y^*)$. The final condition necessary is the following. 
\begin{condition}\label{condition: C6}
    Iterates $x_n$ and $y_n$ are almost surely bounded, i.e.
    \begin{align*}
        \sup_n(\norm{x_n}{} + \norm{y_n}{}) < \infty,\ a.s.
    \end{align*}
\end{condition}
Now that we have all conditions in place, we present the Lemma in \cite{borkar2009stochastic}, guarantying asymptotic tracking of $\{\lambda(y_n) \}$ by trajectory $\{x_n\}$ in the almost surely sense. 

\begin{lemma}[Presented in \cite{borkar2009stochastic}]\label{lemma: convergence2scale}
    Given conditions \ref{condition: C1}, \ref{condition: C2}, \ref{condition: C3}, \ref{condition: C4}, \ref{condition: C5}, \ref{condition: C6}, the iterates converge almost surely to the optimum: 
    \begin{align*}
        (x_n,y_n) \overset{a.s.}{\longrightarrow} \{ (\lambda(y), y)\ :\ y\in \mathbb{R}^k \} 
    \end{align*}
\end{lemma}
\begin{proof}
    The proof is omitted here as it can be found in \cite{borkar2009stochastic}.
\end{proof}

Next, we present the main theorem, found in \cite{BORKAR1997291} or \cite{borkar2009stochastic}.

\begin{theorem}[Convergence of 2 timescale o.d.e. system \cite{BORKAR1997291}]\label{thm: convergence2scale}
    Given conditions \ref{condition: C1}, \ref{condition: C2}, \ref{condition: C3}, \ref{condition: C4}, \ref{condition: C5}, \ref{condition: C6}
    \begin{align*}
        (x_n,y_n) \overset{a.s.}{\longrightarrow} (\lambda(y^*), y^*)
    \end{align*}
\end{theorem}
\begin{proof}
     The proof is omitted here as it can be found in \cite{borkar2009stochastic}.
\end{proof}

\smallskip 

 In the remainder of this section, we show that all of the necessary assumptions hold for the newly defined iterates in Equations~\eqref{eqn: moddedA2CScheme1} and~\eqref{eqn: moddedA2CScheme2}. 
\begin{lemma}[Lipschitz continuity of $\criticgrad,\actorgrad$] \label{lemma: L1}
    Given the analytical gradients $\criticgrad,\actorgrad$ defined above, let $(\criticparam{t},\criticparam{t}')$ and $(\actorparam{t},\actorparam{t}')$ be two sets of critic and actor parameters, respectively. Then,
    \begin{align*}
        |\criticgrad(\actorparam{t},\criticparam{t})-\criticgrad(\actorparam{t}',\criticparam{t}')|&\leq L^{(c)}(|\actorparam{t}-\actorparam{t}'| + |\criticparam{t}-\criticparam{t}'|)\\
        |\actorgrad(\actorparam{t},\criticparam{t})-\actorgrad(\actorparam{t}',\criticparam{t}')| &\leq L^{(a)}(|\actorparam{t}-\actorparam{t}'| + |\criticparam{t}-\criticparam{t}'|)
    \end{align*}
    for suitable Lipschitz constants $L^{(c)},L^{(a)}$.
\end{lemma}
\begin{proof}
    First, we focus on the gradient of the critic $\criticgrad$. To show that $\criticgrad$ is Lipschitz, it is sufficient to show that the partial derivatives $\partial \criticparam{t} \criticgrad$ and $\partial \actorparam{t} \criticgrad$ are continues and bounded. So,
    \begin{align*}
        &\partial \criticparam{t} \criticgrad = \\ &\partial\criticparam{t}\EXP_{\pi_{\actorparam{t}}}\Big[\sum_{k=0}^{K-1}\Big( V_{\criticparam{t}}(S_{t-K+1+k})- \big(\sum_{j=k}^{K-1} \gamma^{j-k} R_{t-K+2+j}\\
        &+\gamma^{K-k} V_{\criticparam{t}}(S_{t+1})\big)\Big)\Big(\phi_{\actorparam{t}}(S_{t+k}) - \gamma^{K-k}\phi_{\actorparam{t}}(S_{t+1}) \Big)\Big]\\
        &= \EXP_{\actorparam{{t}}}\Big[\sum_{k=0}^{K-1}\Big(\phi_{\actorparam{t}}(S_{t+k}) - \gamma^{K-k}\phi_{\actorparam{t}}(S_{t+1})\Big)^2 \Big]
    \end{align*}
    Applying claim \ref{claim: featureVector}, it follows that $\phi_{\actorparam{t}}$ is continuous and bounded.

    Similarly, taking the partial derivative $\partial \actorparam{t} \criticgrad$ gives: 
    \begin{align*}
        &\partial \actorparam{t} \criticgrad =\\
        &\partial \actorparam{{t}}\EXP_{\pi_{\actorparam{t}}}\Big[\sum_{k=0}^{K-1}\Big( V_{\criticparam{t}}(S_{t-K+1+k})- \big(\sum_{j=k}^{K-1} \gamma^{j-k} R_{t-K+2+j}\\
        &+\gamma^{K-k} V_{\criticparam{t}}(S_{t+1})\big)\Big)\Big(\phi_{\actorparam{t}}(S_{t+k}) - \gamma^{K-k}\phi_{\actorparam{t}}(S_{t+1}) \Big)\Big]\\
        &=\partial \actorparam{{t}}\sum_{s}\mu_{\actorparam{t}}(s)\Big[\sum_{k=0}^{K-1}\Big( V_{\criticparam{t}}(s)- \big(\sum_{j=k}^{K-1} \gamma^{j-k} R_{t-K+2+j}\\
        &+\gamma^{K-k} V_{\criticparam{t}}(S_{t+1})\big)\Big)\Big(\phi_{\actorparam{t}}(s) - \gamma^{K-k}\phi_{\actorparam{t}}(S_{t+1}) \Big)\Big]\\\\
        &= \sum_{s\in \St}\partial \actorparam{t}\mu_{\actorparam{t}}(s)\Big[\sum_{k=0}^{K-1}\Big( V_{\criticparam{t}}(s)- \big(\sum_{j=k}^{K-1} \gamma^{j-k} R_{t-K+2+j}\\
        &+\gamma^{K-k} V_{\criticparam{t}}(S_{t+1})\big)\Big)\Big(\phi_{\actorparam{t}}(s) - \gamma^{K-k}\phi_{\actorparam{t}}(S_{t+1}) \Big)\Big]\\
        &  \sum_{s\in \St}\mu_{\actorparam{t}}(s)\Big[\sum_{k=0}^{K-1}\Big( \partial \actorparam{t}V_{\criticparam{t}}(s)\big(\phi_{\actorparam{t}}(s) - \gamma^{K-k}\phi_{\actorparam{t}}(S_{t+1}) \big)\\
        & +V_{\criticparam{t}}(s)\big(\partial \actorparam{t}\phi_{\actorparam{t}}(s) - \gamma^{K-k}\partial \actorparam{t}\phi_{\actorparam{t}}(S_{t+1}) \big)\\
        &- \big(\partial \actorparam{t}\phi_{\actorparam{t}}(s) - \gamma^{K-k}\partial \actorparam{t}\phi_{\actorparam{t}}(S_{t+1}) \big)\sum_{j=k}^{K-1} \gamma^{j-k} R_{t-K+2+j} \\
        &+\gamma^{K-k} \partial \actorparam{t}V_{\criticparam{t}}(S_{t+1})\big(\phi_{\actorparam{t}}(s) - \gamma^{K-k}\phi_{\actorparam{t}}(S_{t+1}) \big)\Big)\Big]\\
        &+\gamma^{K-k} V_{\criticparam{t}}(S_{t+1})\big(\partial \actorparam{t}\phi_{\actorparam{t}}(s) - \gamma^{K-k}\partial \actorparam{t}\phi_{\actorparam{t}}(S_{t+1}) \big)\Big)\Big]
    \end{align*}
    First we focus on the $\sum_{s\in \St}\partial \actorparam{t}\mu_{\actorparam{t}}(s)$ term. The steady state distribution $\mu_{\actorparam{t}}(s)$ depends on the parameter $\actorparam{t}$ through the actor's policy $\pi_{\actorparam{t}}(\cdot| \cdot)$. By Assumption \ref{assumption: boundedPolicyParam}, the derivative of the policy is bounded. Hence $\partial \actorparam{t}Pr(s\longrightarrow s', k, \pi_{\actorparam{t}})$ is bounded for all $k$ and $s\in \St$. It follows that the geometric series $\sum_{k=0}^\infty \gamma^k \partial \actorparam{t}Pr(s\longrightarrow s', k, \pi_{\actorparam{t}})$ converges and therefore $\sum_{s\in \St}\partial \actorparam{t}\mu_{\actorparam{t}}(s) < \infty$. We know that the term $\sum_{k=0}^{K-1}\Big( V_{\criticparam{t}}(s)- \big(\sum_{j=k}^{K-1} \gamma^{j-k} R_{t-K+2+j}+\gamma^{K-k} V_{\criticparam{t}}(S_{t+1})\big)\Big)\Big(\phi_{\actorparam{t}}(s) - \gamma^{K-k}\phi_{\actorparam{t}}(S_{t+1}) \Big)$ is bounded as well, becuase the rewards, feature vectors are bounded, state space is finite and the critic parameters are bounded. Finally, consider the terms $\partial \actorparam{t}\phi_{\criticparam{t}}$ and $\partial \actorparam{t}V_{\criticparam{t}}$. According to Assumption \ref{assumption: boundedPolicyParam} and Claim \ref{claim: featureVector}, the derivative of the feature vectors is bounded. Therefore, it follows that both $\partial \actorparam{t}\phi_{\criticparam{t}}$ and $\partial \actorparam{t}V_{\criticparam{t}}$ are bounded. 
    Therefore, we conclude that there exists a bound $B_{\partial \actorparam{} \criticgrad}$ such that 
    \begin{align*}
        |\partial \actorparam{t} \criticgrad| < B_{\partial \actorparam{} \criticgrad},\ \text{ for all } t.
    \end{align*}
    The continuity of the derivative of the gradient follows from the fact that all terms in the gradient are continuous.
     Next we show that given that the partial derivatives of the gradient are continuous and bounded, we have that the gradient itself is Lipschitz continues. Define 
    \begin{align*}
        S_1 &= \sup_{\criticparam{t}}\{ |\partial \criticparam{t} \criticgrad(\actorparam{t}, \criticparam{t})| \}\\
        S_2 &= \sup_{\actorparam{t}}\{|\partial \actorparam{t} \criticgrad(\actorparam{t}, \criticparam{t})|\}.
    \end{align*}
    Pick two sets of parameters $\actorparam{t}, \criticparam{t}$ and $\actorparam{t}', \criticparam{t}'$. Then 
    \begin{align*}
        \criticgrad(\actorparam{t}, \criticparam{t}) - \criticgrad(\actorparam{t}', \criticparam{t}') &=  \criticgrad(\actorparam{t}, \criticparam{t})-  \criticgrad(\actorparam{t}, \criticparam{t}')\\
        &+ \criticgrad(\actorparam{t}, \criticparam{t}') - \criticgrad(\actorparam{t}', \criticparam{t}')
    \end{align*}
    The triangle inequality implies: 
    \begin{align*}
        |\criticgrad(\actorparam{t}, \criticparam{t}) - \criticgrad(\actorparam{t}', \criticparam{t}')|&\leq |\criticgrad(\actorparam{t}, \criticparam{t})-  \criticgrad(\actorparam{t}, \criticparam{t}')|\\
        &+ |\criticgrad(\actorparam{t}, \criticparam{t}') - \criticgrad(\actorparam{t}', \criticparam{t}')|. 
    \end{align*}
    Then, since $\criticgrad$ is continuous and  the parameter spaces is bounded, by Assumptions \ref{assumption: boundedPolicyParam}, and \ref{assumption: boundedCriticParam},  applying the Mean Value Theorem provides the existence of constants $c_1$ and $c_2$ such that 
    \begin{align*}
        \frac{\criticgrad(\actorparam{t}, \criticparam{t})-  \criticgrad(\actorparam{t}, \criticparam{t}')}{ \criticparam{t} -  \criticparam{t}'} &= \partial \criticparam{t} \criticgrad(\actorparam{t}, c_1) \leq S_1\\
        \frac{\criticgrad(\actorparam{t}, \criticparam{t}') - \criticgrad(\actorparam{t}', \criticparam{t}')}{ \actorparam{t} -  \actorparam{t}'} &= \partial \actorparam{t} \criticgrad(c_2, \criticparam{t}) \leq S_2.
    \end{align*}
    Rearranging the terms in the two expressions above and substituting  them back into the triangle inequality provides: 
    \begin{align*}
        |\criticgrad(\actorparam{t}, \criticparam{t}) - \criticgrad(\actorparam{t}', \criticparam{t}')| &\leq |\criticgrad(\actorparam{t}, \criticparam{t})-  \criticgrad(\actorparam{t}, \criticparam{t}')|\\
        &+ |\criticgrad(\actorparam{t}, \criticparam{t}') - \criticgrad(\actorparam{t}', \criticparam{t}')|\\
        &\leq S_1|\criticparam{t} -  \criticparam{t}'| + S_2|\actorparam{t} -  \actorparam{t}'|.
    \end{align*}
    Let $L^{(c)} = \max\{ S_1, S_2 \}$. It follows that 
    \begin{align*}
        |\criticgrad(\actorparam{t}, \criticparam{t}) - \criticgrad(\actorparam{t}', \criticparam{t}')| &\leq L^{(c)}(|\criticparam{t} -  \criticparam{t}'|+ |\actorparam{t} -  \actorparam{t}'|).
    \end{align*}
    
    Next we repeat the entire process from the derivations above to show that the Lipschitz condition holds for the actor gradient as well. Take the partial derivative with respect to critic parameter of the actor gradient $\actorgrad$ to get 
    \begin{align*}
        &\partial \criticparam{t} \actorgrad \\
        &= \partial \criticparam{t} \EXP_{\pi_{\actorparam{t}}}\Big[\nabla_{\actorparam{t}}\ln\pi_{\actorparam{t}}(A_t|S_t)\\
        &\hphantom{111111111111111}\cdot\sum_{l=0}^{K-1} (\gamma \lambda)^{l}\delta_{\criticparam{t}}(t-K+l+1)\nabla_{\criticparam{t}}\delta_{\criticparam{t}}\Big]\\
        &= \EXP_{\pi_{\actorparam{t}}}\Big[\nabla_{\actorparam{t}}\ln\pi_{\actorparam{t}}(A_t|S_t)\\
        &\hphantom{1111111111111}\cdot\sum_{l=0}^{K-1} (\gamma \lambda)^{l} \Big( \nabla_{\criticparam{t}}\delta_{\criticparam{t}}(t-K+l+1)\nabla_{\criticparam{t}}\delta_{\criticparam{t}}\\
        &\hphantom{1111111111111}+ \delta_{\criticparam{t}}(t-K+l+1)\nabla_{\criticparam{t}}^2\delta_{\criticparam{t}}\Big)\Big]\\
        &= \EXP_{\pi_{\actorparam{t}}}\Big[\nabla_{\actorparam{t}}\ln\pi_{\actorparam{t}}(A_t|S_t)\\
        &\hphantom{1111111111}\cdot\sum_{l=0}^{K-1} (\gamma \lambda)^{l} \Big( \nabla_{\criticparam{t}}\delta_{\criticparam{t}}(t-K+l+1)\nabla_{\criticparam{t}}\delta_{\criticparam{t}}\Big)\Big].
    \end{align*}
    As before, all the terms inside the expectation are continuous and bounded and therefore there exists a constant $B_{\partial \criticparam{} \actorgrad}$ such that 
    \begin{align*}
        |\partial \criticparam{t} \actorgrad(\actorparam{t}, \criticparam{t})| \leq B_{\partial \criticparam{} \actorgrad},\ \text{ for all } t.
    \end{align*}
    Furthermore, taking the partial derivative with respect to the actor parameter $\actorparam{t}$, we get 
    \begin{align*}
        &\partial \actorparam{t} \actorgrad \\
        &= \partial \actorparam{t} \EXP_{\pi_{\actorparam{t}}}\Big[\nabla_{\actorparam{t}}\ln\pi_{\actorparam{t}}(A_t|S_t)\\
        &\hphantom{1111111111111111}\cdot\sum_{l=0}^{K-1} (\gamma \lambda)^{l}\delta_{\criticparam{t}}(t-K+l+1)\nabla_{\criticparam{t}}\delta_{\criticparam{t}}\Big] \\
        &= \partial \actorparam{t}\sum_{s}\mu_{\actorparam{t}}(s)\nabla_{\actorparam{t}}\ln\pi_{\actorparam{t}}(A_t|s)\\
        &\hphantom{1111111111111111}\cdot\sum_{l=0}^{K-1} (\gamma \lambda)^{l}\delta_{\criticparam{t}}(t-K+l+1)\nabla_{\criticparam{t}}\delta_{\criticparam{t}}\\
        &= \sum_{s}\partial \actorparam{t}\mu_{\actorparam{t}}(s)\nabla_{\actorparam{t}}\ln\pi_{\actorparam{t}}(A_t|s)\\
        &\hphantom{1111111111111111}\cdot\sum_{l=0}^{K-1} (\gamma \lambda)^{l}\delta_{\criticparam{t}}(t-K+l+1)\nabla_{\criticparam{t}}\delta_{\criticparam{t}}\\
        &\hphantom{1111111111111111}+ \sum_{s}\mu_{\actorparam{t}}(s)\nabla^2_{\actorparam{t}}\ln\pi_{\actorparam{t}}(A_t|s)\\
        &\hphantom{1111111111111111}\cdot\sum_{l=0}^{K-1} (\gamma \lambda)^{l}\delta_{\criticparam{t}}(t-K+l+1)\nabla_{\criticparam{t}}\delta_{\criticparam{t}}\\
        &\hphantom{1111111111111111}+ \sum_{s}\mu_{\actorparam{t}}(s)\nabla_{\actorparam{t}}\ln\pi_{\actorparam{t}}(A_t|s)\\
        &\hphantom{1111}\cdot\sum_{l=0}^{K-1} (\gamma \lambda)^{l}\Big(\partial \actorparam{t}\delta_{\criticparam{t}}\nabla_{\criticparam{t}}\delta_{\criticparam{t}} + \delta_{\criticparam{t}}\partial \actorparam{t}\nabla_{\criticparam{t}}\delta_{\criticparam{t}} \Big)\Big].
    \end{align*}
    Similarly to the critic's first term, the first term is bounded and continuous. The second term is bounded and continuous due to Assumptions \ref{assumption: boundedPolicyParam} and Claim \ref{claim: boundedGrad} which tells us that $\nabla_{\actorparam{t}}^2\ln\pi_{\actorparam{t}}(A_t|S_t)$ is bounded and continuous. Finally, the third term is bounded and continuous due to Claim \ref{claim: featureVector}. Thus, there exists a constant $B_{\partial \actorparam{} \actorgrad}$ such that 
    \begin{align*}
       |\partial \actorparam{t} \actorgrad(\actorparam{t}, \criticparam{t})| \leq B_{\partial \actorparam{} \actorgrad}.
    \end{align*}
    The rest of the derivation for finding the Lipshitz constant $L^{(a)}$ is along the similar lines as above. 
\end{proof}
Therefore, we showed that Condition \ref{condition: C1} is satisfied. 
We proceed by working on Condition \ref{condition: C6}.
\begin{lemma}[Bounded parameters]\label{lemma:boundedParamSequences}
    The parameters $\criticparam{t}$ and $\actorparam{t}$ are uniformly bounded almost surely,
    \begin{align*}
        \sup_{t}|\criticparam{t}| + \sup_{t}|\actorparam{t}| < \infty,\ a.s.
    \end{align*}
\end{lemma}
\begin{proof}
    We show the derivation for the critic parameter. The derivation for the actor parameter follows the analogous argument. Consider the update scheme: 
    \begin{align*}
        &\criticparam{t+K} = \criticparam{t+K-1}+ \beta(t+K-1) \nonumber \\
    &\cdot \sum_{k=0}^{K-1}\Big( V_{\criticparam{t+K-1}}(S_{t+k})- \big(\sum_{j=k}^{K-1} \gamma^{j-k} R_{t+1+j}(S_{t+j},A_{t+j})  \nonumber\\
    &+\gamma^{K-k} V_{\criticparam{t-K+1}}(S_{t+K})\big)\Big)\nonumber\\
    &\cdot\Big(\phi_{\actorparam{t+K-1}}(S_{t+k}) - \gamma^{K-k}\phi_{\actorparam{t+K-1}}(S_{t+K}) \Big).
    \end{align*}
    From previous derivation for the Lipschitz continuity, it follows that
    \begin{align*}
        |\criticparam{t+K} - \criticparam{t+K-1}| &\leq \beta(t+K-1)B_{\criticgradestim}.
    \end{align*}
    Taking the limit of both sides, as $t\longrightarrow \infty$, we get
    \begin{align*}
        \lim_{t\longrightarrow \infty}|\criticparam{t+K} - \criticparam{t+K-1}| &\leq \lim_{t\longrightarrow \infty}\beta(t+K-1)B_{\criticgradestim} = 0.
    \end{align*}
    Hence the sequence $\{\criticparam{t}\}$ is Cauchy in the complete metric space and thus convergent. All convergent sequences are bounded and so we conclude that  $\{\criticparam{t}\}$ is bounded and hence almost surely bounded. 
    
    An analogous  derivation follows for the sequence  $\{\actorparam{t}\}$.
\end{proof}

Next, we show that the terms $M_{t+1}^{(1)}$ and $M_{t+1}^{(2)}$ are martingale differences with bounded second moments.
\begin{lemma}[Martingale differences]\label{lemma: L2}
     Objects $M_{t+1}^{(c)}$ and $M_{t+1}^{(a)}$ satisfy the martingale difference property and they have bounded second moments.
\end{lemma}
\begin{proof}
    Let $\filtration{t} = \sigma(\criticparam{t'}, \actorparam{t'}, M_{t'}^{(c)},M_{t'}^{(a)}, t' \leq t)$ be a sigma algebra generated by $\criticparam{t'}, \actorparam{t'}, M_{t'}^{(c)},M_{t'}^{(a)}$, such that $t'\leq t$. For the critic, we have that 
    \begin{align*}
        &\EXP[M_{t+1}^{(c)}|\ \filtration{t}]\\
        &=\EXP[\criticgradestim(\actorparam{t},\criticparam{t}) - \criticgrad(\actorparam{t},\criticparam{t})|\ \filtration{t}]\\
        &= \EXP[\criticgradestim(\actorparam{t},\criticparam{t}) |\ \filtration{t}] - \EXP[\criticgrad(\actorparam{t},\criticparam{t})| \filtration{t}]\\
        &= \EXP[\criticgradestim(\actorparam{t},\criticparam{t}) |\ \filtration{t}] \\
        &-\EXP\Big[\EXP_{\pi_{\actorparam{t}}}\Big[\sum_{k=0}^{K-1}\Big( V_{\criticparam{t}}(S_{t-K+1+k})- \big(\sum_{j=k}^{K-1} \gamma^{j-k} R_{t-K+2+j}\\
        &+\gamma^{K-k} V_{\criticparam{t}}(S_{t+1})\big)\Big)\Big(\phi_{\actorparam{t}}(S_{t+k}) - \gamma^{K-k}\phi_{\actorparam{t}}(S_{t+1}) \Big)\Big]\Big| \filtration{t}\Big]\\
        &= \EXP[\criticgradestim(\actorparam{t},\criticparam{t}) |\ \filtration{t}] \\
        &- \EXP\Big[\sum_{k=0}^{K-1}\Big( V_{\criticparam{t}}(s_k)- \big(\sum_{j=k}^{K-1} \gamma^{j-k} R_{t-K+2+j}\\
        &+\gamma^{K-k} V_{\criticparam{t}}(S_{t+1})\big)\Big)\Big(\phi_{\actorparam{t}}(s_{t+k}) - \gamma^{K-k}\phi_{\actorparam{t}}(S_{t+1}) \Big)\Big| \filtration{t}\Big]\\
        & = 0.
    \end{align*}
    The analogous argument is made in the case of actor iterates. 

    Next, we want to show that the second moment is bounded for the critic iterate. Given the derived bounds in Lemma \ref{lemma: L1}, it follows that 
    \begin{align*}
        &\EXP[|M_{t+1}^{(c)}|^2|\ \filtration{t}] \\
        &=  \EXP[|\criticgradestim(\actorparam{t},\criticparam{t}) - \criticgrad(\actorparam{t},\criticparam{t})|^2|\ \filtration{t}] \leq 2\EXP[|\criticgradestim(\actorparam{t},\criticparam{t})|^2 \\
        &\hphantom{1111111111111111111111111111}+| \criticgrad(\actorparam{t},\criticparam{t})|^2|\ \filtration{t}]\\
        &= 2\EXP[|\criticgradestim(\actorparam{t},\criticparam{t})|^2|\ \filtration{t}] + 2\EXP[|\criticgrad(\actorparam{t},\criticparam{t})|^2|\ \filtration{t}]\\
        &\leq 2\EXP[|\criticgradestim(\actorparam{t},\criticparam{t})|^2|\ \filtration{t}] + 2L^{(c)^2}\EXP[(|\actorparam{t}| + |\criticparam{t}|)^2|\filtration{t}]\\
        &\leq 2\EXP[|\criticgradestim(\actorparam{t},\criticparam{t})|^2|\ \filtration{t}] + 4L^{(c)^2}\EXP[|\actorparam{t}|^2 + |\criticparam{t}|^2|\filtration{t}]\\
        &\leq 2B_{\criticgradestim} + 4L^{(c)^2}(|\actorparam{t}|^2 + |\criticparam{t}|^2)
    \end{align*}
    where, as in the lemma \ref{lemma:boundedParamSequences}, the approximate gradient $\criticgradestim(\actorparam{t},\criticparam{t})$ is bounded with bound $B_{\criticgradestim}$. Let $B_{M^{(c)}} := \max\{2B_{\criticgradestim},4L^{(c)^2}\}$. It follows that 
    \begin{align*}
        \EXP[|M_{t+1}^{(c)}|^2|\ \filtration{t}] \leq B_{M^{(c)}} (1 + |\actorparam{t}|^2 + |\criticparam{t}|^2).
    \end{align*}
    An analogous derivation holds for $M_{t+1}^{(a)}$.
\end{proof}
Therefore, condition \ref{condition: C2} is satisfied as well.
\begin{lemma}[Step size choice]
    Denote two real valued, decreasing sequences by $\alpha(t),\beta(t)$, respectively. Then, we define them as 
    \begin{align*}
        \alpha(t) &= \frac{1}{t^{2/3}},\ t=1,2,\dots\\
        \beta(t) &= \frac{1}{t},\ t=1,2,\dots.
 \end{align*}
 It follows that $\alpha(t)$ and $\beta(t)$ satisfy the Robbins-Monro criterion. 
\end{lemma}
\begin{proof}
    Note that for discrete step sizes $t$, it follows that 
    \begin{align*}
        &\sum_{t}  \alpha(t) = \sum_{t} \beta(t) = \infty\\
        &\sum_{t}  \alpha(t)^2 < \infty,\ \sum_{t}  \beta(t)^2 < \infty\ \sum_{t}(\alpha(t)^2+\beta(t)^2) < \infty\\
        &\frac{\beta(t)}{\alpha(t)} = \frac{1}{t^{1/3}} \longrightarrow 0.
    \end{align*}
\end{proof}
Therefore,  Condition \ref{condition: C3} is satisfied as well. 
Next consider the mean ODE, defined as 
\begin{align}\label{eqn: criticODE}
    \dot{\criticparam{}}(t) = \criticgrad(\actorparam{},  \criticparam{}(t))
\end{align}
for a fixed $\actorparam{}$. We want to show that Equation~\eqref{eqn: criticODE} has a unique asymptotically stable equilibrium point $\lambda(\actorparam{})$ where $\lambda$ is a Lipschitz map.
\begin{lemma}[Asymptotic stability of $\dot{\criticparam{}}(t)$]
    Fix the iterate $\actorparam{t}\equiv \actorparam{0}$ and consider the mean ordinary differential equation $\dot{\criticparam{t}}$, defined as 
    \begin{align}
        \dot{\criticparam{}}(t) = \criticgrad(\actorparam{0},  \criticparam{}(t)).
    \end{align}
    It follows that there exists a unique asymptotically stable equilibrium point $\criticparam{}(\actorparam{0})^*$.
\end{lemma}
\begin{proof}
    Because $\criticgrad$ is Lipschitz, it follows that the ODE \eqref{eqn: criticODE} is well posed, i.e. it has a unique solution for each initial condition $(\actorparam{0},\criticparam{}(0))$ and the solution continuously depends on $(\actorparam{0},\criticparam{}(0))$. By  the Picard-Lindel\"of theorem, a unique equilibrium point of the system exists on some interval $I = [t_0 - \epsilon, t_0 + \epsilon]$. Denote such an equilibrium point as $\criticparam{}(\actorparam{0})^*$, and define a corresponding Lypaunov function $\mathcal{L}_{\criticparam{}}(\criticparam{}(t) - \criticparam{}(\actorparam{0})^*)$ as
    \begin{align*}
        \mathcal{L}_{\criticparam{}}(\criticparam{}(t)- \criticparam{}(\actorparam{0})^*):= L_{\criticgrad}(\actorparam{},\criticparam{}(\actorparam{0})^*) - L_{\criticgrad}(\actorparam{},\criticparam{}(t)).
    \end{align*}
    Computing the derivative of $ \mathcal{L}_{\criticparam{}}$ with respect to $t$ provides: 
    \begin{align*}
        &\frac{d}{dt}\mathcal{L}_{\criticparam{}}(\criticparam{}(t)- \criticparam{}(\actorparam{0})^*) = \frac{d}{d\criticparam{}}\mathcal{L}_{\criticparam{}}(\criticparam{}(t)- \criticparam{}(\actorparam{0})^*)\frac{d}{dt}\criticparam{}(t)\\
        &= -\frac{\partial}{\partial \criticparam{}}L_{\criticgrad}(\actorparam{},\criticparam{}(t))\cdot \criticgrad(\actorparam{},  \criticparam{}(t)) = -\criticgrad(\actorparam{},  \criticparam{}(t))^2.\\
    \end{align*}
    Because $L_{\criticgrad}$ is convex, it follows that the Lyapunov function $\mathcal{L}_{\criticparam{}}$ is positive definite. Furthermore, it is continuous differentiable according to Claim \ref{claim: featureVector}. Finally, it satisfies 
    \begin{align*}
        \frac{d}{d\criticparam{}}\mathcal{L}_{\criticparam{}}(\criticparam{}(t)- \criticparam{}(\actorparam{0})^*)\frac{d}{dt}\criticparam{}(t) \leq 0,\ \text{ for all } \criticparam{}(t).
    \end{align*}
    Then, by the Lyapunov theorem on stability, it follows that $\criticparam{}(\actorparam{0})^*$ is a Lyapunov stable equilibrium of our mean ODE. Furthermore, because ODE \eqref{eqn: criticODE} is well posed for any initial solution and it is Lyapunov stable, it follows that ODE \eqref{eqn: criticODE} is (globally) asymptotically stable with the stable equilibrium $\criticparam{}(\actorparam{0})^*$.
\end{proof}

We also need to establish the existence of a Lipschitz map $\lambda$ which maps from the space of parameters $\actorparam{}$ to the space of parameters $\criticparam{}$. In order to this, we follow the work of \cite{PPOconvergence}.
\begin{lemma}[Existence of Lipschitz map]
    There exists a map $\lambda(\theta): \Theta \longrightarrow \Omega$ such that $\lambda(\cdot)$ is Lipschitz.
\end{lemma}
\begin{proof}
    We proceed by applying the implicit function theorem (IFT) to the critic's gradient and establish that there exists a continuously differentiable function $\lambda(\cdot)$. Consider function $f$ defined as
    \begin{align*}
        f(\actorparam{}, \cdot):= \nabla_{\criticparam{}} L_{\criticgrad}(\actorparam{},\cdot).
    \end{align*}
    Let $\criticparam{}(\actorparam{0})^*$ be the critical point of function $f$ such that 
    \begin{align*}
        f(\actorparam{0}, \criticparam{}(\actorparam{0})^*)= \nabla_{\criticparam{}} L_{\criticgrad}(\actorparam{0},\criticparam{}(\actorparam{0})^*) = 0.
    \end{align*}
    Let $(\actorparam{0},\criticparam{0})$ be the initial parameter pair and define a neighbourhood $U_{\actorparam{0}}(\criticparam{0})$ that connects $\criticparam{0}$ with the local minimum $\criticparam{}(\actorparam{0})^*$ of function $f$ such that there are no other local minima in this neighbourhood. This can be done by defining a neighbourhood as a narrow strip on the loss surface of function $L_{\criticgrad}$ which starts around $\criticparam{0}$ and extends towards the local minima $\criticparam{}(\actorparam{0})^*$.
    Then by the implicit function theorem (IFT), it follows that there exists a neighbourhood $V(\actorparam{0})$ around initial point $\actorparam{0}$ and continuously differentiable function $\lambda: V\longrightarrow U_{\actorparam{}}$ such that 
    \begin{align*}
        f(\actorparam{0}, \lambda(\actorparam{}))= \nabla_{\criticparam{}} L_{\criticgrad}(\actorparam{0},\criticparam{}(\actorparam{0})^*) = 0,\ \actorparam{}\in V(\actorparam{0}).
    \end{align*}
    where
    \begin{align*}
        \lambda(\actorparam{}) = \criticparam{}(\actorparam{0})^*,\ \actorparam{}\in V(\actorparam{0}).
    \end{align*}
    Because the parameter spaces $\Theta$ and $\Omega$ are bounded, it follows that by Extreme Value theorem, that function $\lambda$ is bounded. Furthermore, because $\lambda$ is continues, differentiable and bounded, it follows that by the Mean Value theorem that $\lambda$ is Lipschitz. 
\end{proof}
Now that we showed that condition \ref{condition: C1} is satisfied, it follows that by applying lemma \ref{lemma: convergence2scale}, we establish convergence of the system
\begin{align*}
    \dot{\criticparam{}}(t) &= \criticgrad(\criticparam{}(t), \actorparam{}(t))\\
    \dot{\actorparam{}}(t) &= 0
\end{align*}
which implies that 
\begin{align}
    (\criticparam{t},\actorparam{t}) \overset{a.s.}{\longrightarrow} \{ (\lambda(\actorparam{}), \actorparam{})\ :\ \actorparam{}\in \mathbb{R}^n \}. 
\end{align}

The last condition we need to check is Condition \ref{condition: C5}. In this case, we cannot apply the Lyapunov technique again as the policy gradient not need be a positive definite function. However, we consider the following well established result about gradient systems.
\begin{lemma}[Convergence of gradient trajectory \cite{AbsilKurdykaGradientSystm}]\label{lemma: gradSystemConvergence}
    Assume that $f$ is a $C^2$ function and let $x(t)$ be a solution trajectory of the gradient system 
    \begin{align*}
        \dot{x}(t) = -\nabla f(x(t))
    \end{align*}
    contained in a compact set $K\subseteq \mathbb{R}^n$. Then $x(t)$ approaches critical set $C_K = \{ y\in K\ :\ \nabla f(y) = 0 \}$ as $t$ approaches infinity. That is, $\lim_{t\longrightarrow \infty} \norm{x(t) - y}{} = 0$.
\end{lemma}
\begin{proof}
    The proof is shown by using Theorem 9.4.4 in \cite{hirsch1974differential} and Lemma 3.1 in \cite{khalil2002nonlinear}.
\end{proof}
\begin{lemma}[Asymptotic stability of $\dot{\actorparam{}}(t)$]
    Consider the mean ordinary differential equation $\dot{\actorparam{}}(t)$, defined as 
    \begin{align}
        \dot{\actorparam{}}(t) = \actorgrad\big(\actorparam{}(t), \lambda(\actorparam{}(t))\big).
    \end{align}
    It follows that there exists a unique asymptotically stable equilibrium point $\big(\actorparam{}^*, \lambda(\actorparam{}^*)\big)$.
\end{lemma}
\begin{proof}
    In the case of the ODE $ \dot{\actorparam{}}(t)$ above, the dynamics from Lemma \ref{lemma: gradSystemConvergence} is the negative of the actor's loss function, defined as $-f(\actorparam{}(t)) := L_{a}(\actorparam{}(t))$. Furthermore, we approximation the gradient with the projected gradient $\actorgrad(\actorparam{}(t),\lambda(\actorparam{}(t)))$, as the projection error is orthogonal to the vector space $\Psi_{\actorparam{}}$, and get
    \begin{align*}
        \actorgrad(\actorparam{}(t),\lambda(\actorparam{}(t))) \approx
         \nabla L_{a}(\actorparam{}(t)) = -\nabla f(\actorparam{}(t)).
    \end{align*}
     Hence, we consider the system
    \begin{align*}
        \dot{\actorparam{}}(t) = \actorgrad\big(\actorparam{}(t), \lambda(\actorparam{}(t))\big).
    \end{align*}
    According to claim \ref{claim: boundedGrad}, $L_{a}$ is twice continuously differentiable with respect to actor parameter $\actorparam{}$. Furthermore, the trajectory $\big(\actorparam{}(t), \lambda(\actorparam{}(t))\big)$ remains in the compact space $K$ due to the parameter space $\Omega \times \Theta$ being closed and bounded. Hence, by applying lemma \ref{lemma: gradSystemConvergence}, it follows that 
    \begin{align*}
        \big(\actorparam{}(t), \lambda(\actorparam{}(t))\big) \overset{a.s.}{\longrightarrow} \big(\actorparam{}^*, \lambda(\actorparam{}^*)\big).
    \end{align*}
\end{proof}
Finally, due to conditions \ref{condition: C1}, \ref{condition: C2}, \ref{condition: C3}, \ref{condition: C4}, \ref{condition: C5}, \ref{condition: C6} being satisfied, we can apply Lemma \ref{lemma: convergence2scale} and Theorem \ref{thm: convergence2scale} to show that the  Actor-Critic algorithm converges to its approximate optimal policy $\pi^*$.

This completes the proof of Theorem~\ref{thm: main}. \label{last page of proof}


\section{Further details on the experiments} \label{appendix:experimentData}

The purpose of this section is to provide additional details on the computing resources used and how we obtained or generated the real and the synthetic data used in Section 5.  

\paragraph{Computing resources and setup.}
The sampling procedure is done using a standard CPU offered by Google Colab, although we note that some speed up might be possible by leveraging GPU functionality.  
For the $100$-node graph, with edge probability of $p=0.3$, which corresponds to fiber point $g\in \mathbb{Z}^{4950}$, the Actor-Critic Sampler Algorithm~\ref{alg:A2CMoveGen} approximately takes $4.5$ minutes to generate 150 feasible moves.

\paragraph{Data generation: networks.} 
We test the algorithm on a co-authorship dataset and on synthetic network data. The co-authorship data set was  published in \cite{JiJinAuthorsAOAS}. 
Since this is a  sparse network of a very particular structure (see Figure~\ref{fig: chi_square_dist_p_val}), we also test our algorithm on families of randomly generated graphs. Here we show results obtained by generating  Erd\"{o}s-R\'enyi   graphs with edge probability parameter $p$. 
  We use \url{https://networkx.org} package in python for this step. We choose these types of networks because they are easy to generate and allow sparsity control in the data, in order to  gain a better understanding of algorithm robustness. 
The synthetic datasets do not fit the model used for the co-authorship data; this is expected and we do not report those results as they are not statistically interesting. Rather, we use the random graphs to test out how well the \emph{learning} works on graphs of varying sparsity. 

\paragraph{Contingency table data.} 

Section 5, part (2), describes the results of testing two types of data sets that appear in the literature. 

The first type is  a $4\times4$ table of individuals cross-classified according to salary range and job satisfaction. In \cite{MB25years}, there are two tables: one that fits the model of independence and one that does not. Lookin into the details of the model that is supposed to be rejected, \cite{MB25years} illustrated that the Markov chain Metropolis-Hastings algorithm for sampling the fiber provides very unreliable results, sometimes rejecting and sometimes accepting the model of independence, in repeated sampling. In contrast, our algorithm performed consistently: always rejecting. 

\begin{figure}[H]
    \centering
    \begin{tabular}{cc}
    \subf{\includegraphics[width=75mm]{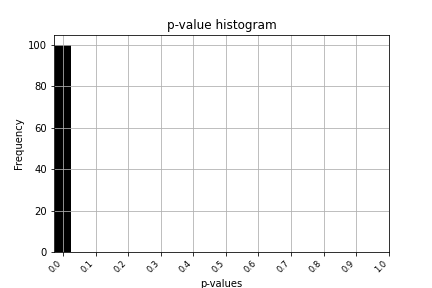}}
         {}
    \end{tabular}
    \caption{Histograms for 100 p-values from 100 exchangeable samples of size 100, obtained through simulation of 100 independent Markov chains using the trained optimal policy $\pi^*$. In this case, we used the ``bad" lattice basis and showed consistent results, empirically proving that our algorithm can work with any lattice basis. Compare to \cite[Figure 2]{MB25years}.
    }  
    \label{fig: p_value_test_bad_basis}
\end{figure}

Here we note one possible point of contention: \emph{the lattice basis we use and the lattice basis \cite{MB25years} use are not the same}. This is because we use {\tt python}'s linear algebra routines to compute a lattice bases, while they used a carefully constructed lattice basis. To that end, we re-ran our experiments to force our algorithm to use the same ``bad" lattice basis as they use. The results are summarized in Figure~\ref{fig: p_value_test_bad_basis}: the Actor Critic Sampler still performs consistently in repeated runs, discovering the fiber of this extreme sparse table in a way that is comparable to a Markov basis Metropolis-Hastings algorithm. 


\section{Exploration of failure}\label{appendix:exploring failure}

The examples we included in the main body of the paper illustrate how the algorithm can succeed. In this section, we summarize scenarios where it is not straightforward to use the algorithm. While using a lattice basis (rather than Markov or Gr\"obner) already alleviates the computational challenges for forming the bases elements for exploring the fiber, there exist scenarios where even the lattice basis computation appears challenging. 

For a concrete example, the full coauthorship data set, of which data set (1) in Section~\ref{sec: simulations} is a subgraph, corresponds to adjacency vectors of size the order of $10$ million. This size alone choked the standard linear algebra algorithms trying to compute -- and let alone store and combine -- the lattice basis for the vector space $\ker M$.  Therefore,  we used some intuitive graph techniques to decompose the graph into the most highly connected components, using various graph metrics.  The $23$-node subgraph  an example of such a meaningful component. To analyze the full-sized data set, further  decomposition would be required. This additional scaling challenge is the reason we propose the optional additional step to decompose the initial solution and allow the agent to learn on sub-problems.

The decomposition step further alleviates the computational burden of producing moves for large matrices $M$.  We can demonstrate how such a decomposition can be obtained and provide pseudocode tor how to re-combine moves tor larger data sets in Appendix~\ref{appendix: pseudocode}. 
 Corollary~\ref{cor: reconstruction} ensures that optimality of the policy will be preserved under such decompositions.

Following Corollary~\ref{cor: reconstruction}, we state that the exact way in which a problem subdivision is achieved may depend on the particular problem and data structure. Our methodology does require a decomposition strategy to be applied to problems with encoding vectors of size order of a million, since otherwise even a lattice basis may not be easily computable. That's a linear algebra scaling problem. Beyond the current paper, future work may be able to address this using a sparse encoding of matrices and bases.


\section{Convergence in practice}
\begin{figure}[H]
    \centering
\includegraphics[scale=.55]{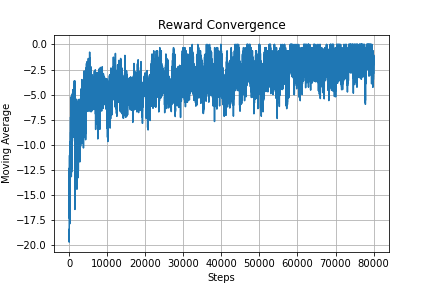}
 \caption{Moving average reward collected over $80,000$ steps.
    }  
    \label{fig: convergence}
\end{figure}

Based on our problem formulation, an optimal policy is one that produces feasible moves and continually samples from the fiber. To determine if the algorithm has converged to the optimal policy, one considers the reward.
We plot the moving average reward for a fiber sampling problem in Figure~\ref{fig: convergence}. As one can observe, the reward is converging towards constant $0$, which means that the agent is producing more and more feasible actions and at the same time less frequently violating the feasibility condition. This is exactly the desired behavior of the optimal policy for the fiber sampling problem. Because the agent is sampling the action from a distribution, it will occasionally make mistakes and step outside the positive quadrant  of the lattice, hence the negative rewards. However, in our application, the randomness is desirable due to the fact that we are looking for a random sample of the data.  Thus when the agent samples a unfeasible move, we simply throw it away and sample again.

\section{Pseudo code for Actor-Critic Sampler} \label{appendix: pseudocode} 

If the problem is too large for even the lattice basis computation, we can decompose the original data (graph) into subproblems or subgraphs and work on them separatly. To achieve this, we use the following algorithm. Given an initial point $G$ (e.g. $G$ is a graph), Algorithm \ref{alg:initialDecomp} decomposes $G$ into sub-points for which we can compute the lattice basis. In the case of graph data, this decomposition can be done using $k$-core decomposition, bridge cuts or induced subgraphs.
\begin{algorithm}[H]
\caption{{Initial Point Decomposition}}
\label{alg:initialDecomp}
\hspace*{\algorithmicindent} \textbf{Input:} Initial point (e.g. Graph $G = (E,V)$)\\
\hspace*{\algorithmicindent} \textbf{Output:} Collection of sub-initial solutions $\mathcal{G}$, $\mathcal{B}_{\mathcal{G}}$ - lattice basis for each sub-problem in $\mathcal{G}$
\begin{algorithmic}[1]
\State Use k-core decomposition, bridge cuts, induced subgraphs to obtain $\mathcal{G}$
\For{$g\in \mathcal{G}$}
\State Compute sufficient statistic $Ag = b_g$, lattice basis $\mathcal{B}_{g}$
\EndFor
\end{algorithmic}
\end{algorithm}

After obtaining sub-problems and their corresponding lattice basis, we employ the A2C algorithm learn the optimal sampling policy for each subgraph.

\begin{algorithm}[H]
\caption{{Pseudo code of A2C Sampler}}
\label{alg:A2CMoveGen}
\hspace*{\algorithmicindent} \textbf{Input:} Learning rates $\alpha,\beta$, Collection of subgraphs $\mathcal{G}$, Lattice basis $\mathcal{B}_{\mathcal{G}}$, Size of roll-out $K$, Total number of steps $t_{max}$\\
\hspace*{\algorithmicindent} \textbf{Output:} $\pi^*$ OR/AND set of feasible moves $\mathcal{M}$
\begin{algorithmic}[1]
\Repeat
\State Reset gradients: $d\actorparam{} \leftarrow 0\ ,  d\criticparam{} \leftarrow 0$
\State $t_{start} \leftarrow t$
\State Get state $S_t = s$
\Repeat
\State Preform $A_t$ according to policy $\pi(A_t|S_t; \actorparam{})$
\State Receive reward $R_{t+1}$ and new state $S_{t+1}$
\State $t\leftarrow t+1$
\Until {terminal $S_t$ \textbf{or} $t-t_{start} == K$}
\State Compute temporal differences for each time step during roll-out
\begin{align*}
    &\delta_{\criticparam{t+K-1}}(t)\\
    &= R_{t+1}(S_{t},A_{t}) + \gamma V_{\criticparam{t+K-1}}(S_{t}+A_t) - V_{\criticparam{t+K-1}}(S_t),\\ &\hphantom{111111111111111111111}t=t_{start},\dots,K
\end{align*}
\State Accumulate gradients w.r.t. $\actorparam{}$ and update parameter based on actor update \ref{def: actorUpdate}.
\State Accumulate gradients w.r.t. $\criticparam{}$ and update parameter based on critic update \ref{def: criticUpdate}.

\Until{$t>t_{max}$}

\end{algorithmic}
\end{algorithm}

With the learned policy, we apply Algorithm \ref{alg:permutation} to the optimal moves, in order to lift them into a higher dimensional space and sample from the original fiber. Each sub-problem and its generated moves hold information about which sub-structure they belong to. We track the edge labels of the sub-moves and compare them to the edge labels of the original problem. Our lifted move starts as an empty list which iteratively gets populated with integer values. Whenever we have a match between edges from the the sub-move and original problem, we append the value of the sub-problem at that position to the empty list. If there is no such match, we append a zero. In the end, we obtain a move with size matching the original problem. Because we are padding the sub-move with zeros, such move is still applicable to the original problem.

\begin{algorithm}[H]
\caption{{Move Permutation}}
\label{alg:permutation}
\hspace*{\algorithmicindent} \textbf{Input:} Feasible moves $\mathcal{M}$, list of lexicographically ordered edges $\mathcal{E}$ of fully connected graph\\
\hspace*{\algorithmicindent} \textbf{Output:} Permuted moves $\mathcal{M}^*$
\begin{algorithmic}[1]
\For{$m\in \mathcal{M}$}
\State $m^* \leftarrow [\ ]$ 
\For{$e \text{ in } \mathcal{E}$}
\If{$e \text{ not in } m$}
\State $m^*.\text{append}(0)$
\Else{}
\State $m^*.\text{append}(m[indx(e)])$
\EndIf
\EndFor
\EndFor
\end{algorithmic}
\end{algorithm}



\end{document}